\documentclass[10pt]{article}
\usepackage[accepted]{tmlr}

\usepackage{amsmath,amsfonts,amssymb}
\usepackage{array}
\usepackage{afterpage}
\usepackage{booktabs}
\usepackage{colortbl}
\usepackage{float}
\usepackage{graphicx}
\usepackage{longtable}
\usepackage{makecell}
\usepackage{multirow}
\usepackage{nicefrac}
\usepackage{pifont}
\usepackage{ragged2e}
\usepackage{textcomp}
\usepackage{url}
\let\DefaultUrlBreaks\UrlBreaks
\usepackage{xurl}
\let\BibliographyUrlBreaks\UrlBreaks
\let\UrlBreaks\DefaultUrlBreaks
\usepackage{wrapfig}
\usepackage{xcolor}
\usepackage{xspace}
\usepackage{hyperref}

\DeclareRobustCommand{\ourbench}{{\normalfont\textsc{UniGenBench++}}}

\newcommand{\ie}{\emph{i.e., }}
\newcommand{\eg}{\emph{e.g., }}

\hypersetup{
    colorlinks=true,
    linkcolor=red,
    citecolor=blue,
    urlcolor=magenta
}

\title{UniGenBench++: A Unified Semantic Evaluation Benchmark for Text-to-Image Generation}

\author{%
\name Yibin Wang\textsuperscript{*\textdaggerdbl} \email yibinwang1121@163.com \\
\addr Fudan University; Shanghai Innovation Institute; Hunyuan, Tencent
\AND
\name Zhimin Li\textsuperscript{*} \email zhiminli.cn@outlook.com \\
\addr Hunyuan, Tencent
\AND
\name Yuhang Zang\textsuperscript{*} \email zang0012@ntu.edu.sg \\
\addr Shanghai AI Laboratory
\AND
\name Jiazi Bu \email bujiazi001@sjtu.edu.cn \\
\addr Shanghai Jiao Tong University; Shanghai AI Laboratory
\AND
\name Yujie Zhou \email yujieouo@sjtu.edu.cn \\
\addr Shanghai Jiao Tong University; Shanghai AI Laboratory
\AND
\name Yi Xin \email xinyi@smail.nju.edu.cn \\
\addr Shanghai Innovation Institute
\AND
\name Junjun He \email hejunjun@sjtu.edu.cn \\
\addr Shanghai Innovation Institute; Shanghai AI Laboratory
\AND
\name Chunyu Wang\textsuperscript{\textdagger} \email chunyu.wangdlut@gmail.com \\
\addr Hunyuan, Tencent
\AND
\name Qinglin Lu\textsuperscript{\textdagger} \email qinglinlu@tencent.com \\
\addr Hunyuan, Tencent
\AND
\name Cheng Jin\textsuperscript{\textdagger} \email jc@fudan.edu.cn \\
\addr Fudan University; Shanghai Innovation Institute
\AND
\name Jiaqi Wang\textsuperscript{\textdagger} \email wjqdev@gmail.com \\
\addr Shanghai Innovation Institute
}

\def\month{11}
\def\year{2026}
\def\openreview{%
\url{https://openreview.net/forum?id=Id3q3eJw4M}\\%
{\bf Project Page:} \url{https://codegoat24.github.io/UniGenBench}%
}

\begin{document}

\maketitle

\begin{NoHyper}
\begingroup
\renewcommand{\thefootnote}{}
\footnotetext{\textsuperscript{*} Equal contribution.\quad \textsuperscript{\textdagger} Corresponding authors.\quad \textsuperscript{\textdaggerdbl} Project lead.}
\footnotetext{Large language models were used to assist with language editing.}
\endgroup
\end{NoHyper}

% Keep the abstract together after the long author list without shrinking the template.
\clearpage

\begin{abstract}
% !TEX root = main.tex

Recent progress in text-to-image (T2I) generation underscores the importance of reliable benchmarks in evaluating how accurately generated images reflect the semantics of their textual prompt. However, (1) existing benchmarks lack the diversity of prompt scenarios and {coverage beyond English}, both essential for real-world applicability; (2) they offer only coarse evaluations across primary dimensions, covering a narrow range of sub-dimensions, and fall short in fine-grained sub-dimension assessment.  To address these limitations, we introduce \ourbench, a unified semantic assessment benchmark for T2I generation. Specifically, it comprises 600 prompts organized hierarchically to ensure both coverage and efficiency: (1) it spans across diverse real-world scenarios, \ie 5 main prompt themes and 20 subthemes; (2) comprehensively probes T2I models' semantic consistency over 10 primary and 27 sub evaluation criteria, with each prompt assessing multiple test points. To rigorously assess model robustness to variations in language and prompt length, we provide both English and Chinese versions of each prompt in short and long forms.
Leveraging general world knowledge and fine-grained image understanding capabilities of a closed-source Multi-modal Large Language Model (MLLM), \ie Gemini-2.5-Pro, we develop an effective pipeline for reliable benchmark construction and streamlined model assessment. Moreover, to further facilitate community use, we train a robust evaluation model that enables offline assessment of T2I generation. Through comprehensive benchmarking of open- and closed-source models, we systematically reveal their strengths and weaknesses across various aspects.

\end{abstract}


\begin{figure}[ht]
\centering
\includegraphics[width=\textwidth]{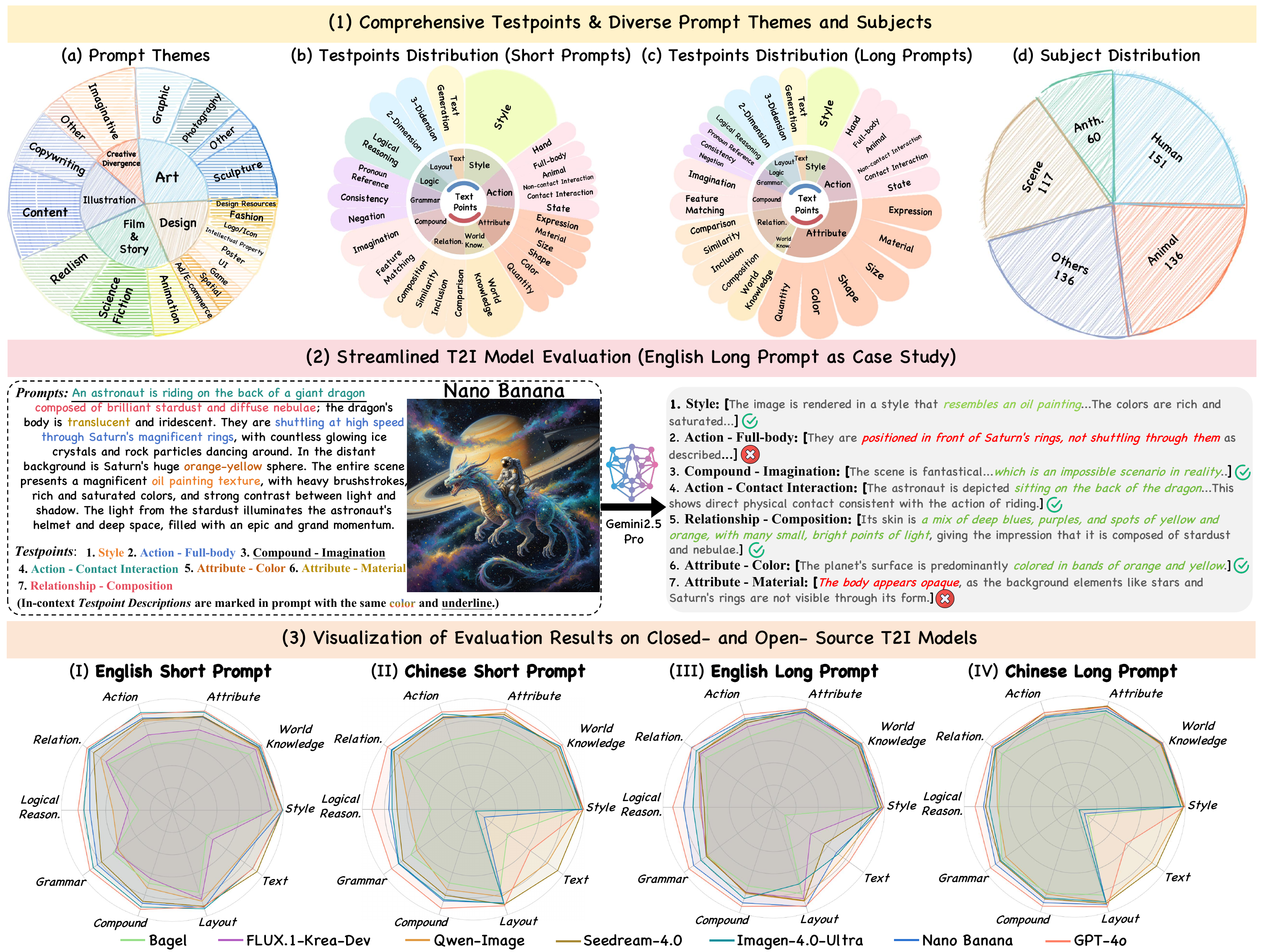}
\caption{\textbf{Benchmark Overview.} (1) Our \ourbench\ covers diverse prompt themes, subjects, and comprehensive evaluation criteria. (2) Each prompt includes multiple test points and is assessed through a streamlined MLLM-based pipeline for reliable and efficient evaluation. (3) We conduct comprehensive evaluations of both open- and closed-source models using both English and Chinese prompts in short and long forms, systematically revealing their strengths and weaknesses across various aspects.}
\label{fig:teaser}
\end{figure}

% !TEX root = main.tex
\section{Introduction}
Recent progress in text-to-image (T2I) generation \citep{diffusion,sd,liu2022flow,dall-e,sdxl,esser2024scaling,show-o2,omnigen2,flux,seedream3,seedream4,cai2025hidream,nano-banana,gpt-image-1,qwen-image,plaground25,DreamText,imagen4,flux-krea} has highlighted the ability to generate high-quality images directly from natural language descriptions. Technically, current T2I models can be broadly divided into two paradigms. (1) Diffusion-based methods, including Stable Diffusion \citep{sd,sdxl}, Playground \citep{plaground25}, and FLUX \citep{flux,flux-krea}, iteratively refine Gaussian noise using U-Net or Transformer backbones to generate images. (2) Autoregressive (AR) approaches, such as Infinity \citep{han2025infinity}, Janus series \citep{janus,janus-pro,ma2024janusflow}, and BLIP3-o \citep{blip3o}, treat images as token sequences and synthesize them via next-token prediction or progressive scaling. Recent methods incorporate reinforcement learning \citep{dpo,guo2025deepseek,unifiedreward-think,LiFT} to improve T2I models' instruction following capability \citep{prefgrpo,tong2025delving} and the visual quality of generated images \citep{flowgrpo,xue2025dancegrpo,zhou2025text}.
With these rapid advancements, assessing T2I models, particularly their semantic generation capability, \ie \textit{how accurately generated images reflect the semantics of their textual prompt}, has emerged as a critical challenge. Traditional benchmarks \citep{ghosh2023geneval,t2i-compbench} typically evaluate T2I models by probing various compositional generation and employ CLIP-based metrics for quantitative assessment. However, CLIP-based scorers remain limited in capturing the fine-grained semantic information and complex world knowledge or logical reasoning. Therefore, several studies \citep{niu2025wise,t2i-reasonbench} evaluate the implicit semantic understanding and world knowledge integration capabilities of T2I models using powerful visual-language models (VLMs) \citep{gpt4o} as the evaluator. Recent efforts broaden T2I evaluation by incorporating long-prompt semantics generation \citep{dpg-bench} and additional evaluation dimensions \citep{wei2025tiif} such as style and text generation.
\afterpage{\afterpage{% !TEX root = ../main.tex
\begin{table}[t]
\setlength{\tabcolsep}{7.5pt}
\centering
% \scriptsize
\caption{\textbf{Semantic Evaluation Benchmark Comparison.} ``-'' indicates that the aspect is not discussed in its original paper.}

\resizebox{\textwidth}{!}{%
\begin{tabular}{lccccccccc}
\toprule
Benchmark & \makecell[c]{Primary\\ Dimension} & \makecell[c]{Sub\\ Dimension} & \makecell[c]{Prompt\\ Theme} &  \makecell[c]{Prompt\\ Length} &  \makecell[c]{Prompt\\ Num.} & \makecell[c]{Multi-Testpoint\\ per Prompt} & \makecell[c]{Multilingual\\ Support} & \makecell[c]{Dedicated Offline\\Eval Model}\\ \midrule
GenEval  & 6 & - & - & short &  553 & \ding{55}  & \ding{55} & \checkmark\\ 
T2I-CompBench++ & 8 & -& - & short &   2,400 & \ding{55}  & \ding{55} & \checkmark\\ 
DPG-Bench & 5 & - & - & long &   1,065 & \ding{55} & \ding{55} & \ding{55}\\ 
WISE & 6 & - & - & short &  1,000 & \ding{55} &\ding{55} & \ding{55}\\ 
TIIF-Bench  & 9 & - & - & short/long &  5,000 & 1$\sim$2 &\ding{55} & \ding{55}\\ 

\hline
UniGenBench++ (\textbf{Ours}) & {10} & 27 & {20} & short/long &  600 & 1$\sim$10 &\checkmark &  \checkmark \\	\bottomrule
\end{tabular}
}

\label{tab:bench_comparison}
% \vspace{-0.4cm}
\end{table}
}}

Despite effectiveness, as shown in Tab.~\ref{tab:bench_comparison}, these benchmarks encounter two key limitations: (1) \textbf{Coarse evaluation on limited dimensions:} they cover limited general dimensions (\eg lacking \textit{grammar}, \textit{action}), within which the sub-dimension coverage is also limited (\eg lacking \textit{relation}-\textit{similarity}, \textit{inclusion}), and are incapable of fine-grained assessment for each sub-dimension;
(2) {\textbf{Limited prompt-scenario diversity and language coverage:} existing benchmarks often emphasize evaluation dimensions, with limited coverage of diverse application scenarios and evaluation beyond English.}

In light of these challenges, this work posits that \textbf{(1)} existing T2I models have already shown strong performance on several primary dimensions (\eg attributes) in current benchmarks \citep{ghosh2023geneval,dpg-bench,wei2025tiif}. This highlights the necessity of further decomposing these dimensions into explicit, comprehensive sub-dimension-level test points (\eg attribute-expression) to enable a more comprehensive and diagnostic evaluation of model capabilities, thereby uncovering fine-grained weaknesses that coarse metrics often overlook. \textbf{(2)} Real-world T2I generation involves diverse scenarios (\eg UI design, graphic art) and naturally spans multiple languages. The absence of such diversity in current benchmarks limits evaluation robustness, causing models that excel in constrained settings to falter in real-world applications.

To this end, we introduce \ourbench, a unified semantic-generation benchmark tailored for fine-grained and comprehensive evaluation of T2I models. As illustrated in Fig.~\ref{fig:teaser} (1), this benchmark \textbf{comprises 600 prompts organized within a hierarchical structure that ensures both coverage and efficiency}:
(i) It provides a comprehensive assessment of semantic consistency across 10 primary and 27 sub-dimensions, each prompt targeting multiple specific test points. This design strikes a balance between fine-grained evaluation and efficiency, ensuring the benchmark captures diverse aspects of model semantic generation capability.
(ii) It spans 5 major real-world primary generation scenarios and 20 sub-scenarios with diverse subject categories, covering application-relevant domains and enabling structured evaluation of semantic requirements across diverse generation scenarios. Furthermore, to enable systematic evaluation of {models'} sensitivity to language and prompt length, each prompt is provided in both English and Chinese, and in short and long forms.
For effective and efficient evaluation, in contrast to widely adopted paradigms, such as multi-turn conversational assessments with VLMs for each image evaluation \citep{wei2025tiif,t2i-compbench,ghosh2023geneval}, our benchmark \textbf{introduces a streamlined, point-wise evaluation pipeline}, as illustrated in Fig.~\ref{fig:teaser} (2): given a prompt, its corresponding image, and a set of explicitly designed test points (each accompanied by its in-context description within the prompt), the evaluation model, \ie Gemini-2.5-Pro \citep{gemini2.5pro}, sequentially analyses whether each semantic requirement is faithfully represented in the image and assigns an appropriate score.
This lightweight and structured design reduces evaluation complexity while ensuring consistent, fine-grained, and interpretable judgments for every test point, thereby enabling more efficient and diagnostic assessment of T2I models.
Moreover, to further facilitate community use, we provide a robust evaluation model that supports offline assessment of T2I generation.

We conduct a comprehensive bilingual (English/Chinese) and length-varied (short/long prompt) benchmarking across both closed-source models, such as GPT-4o \citep{gpt-image-1}, Nano Banana Pro \citep{nano-banana}, Seedream-4.5 \citep{seedream4}, and FLUX-Kontext-Max \citep{flux}, as well as leading open-source counterparts, including Z-Image \citep{z_image}, Qwen-Image \citep{qwen-image}, FLUX.2-dev \citep{flux2}, Lumina-DiMOO \citep{lumina_dimoo} and Bagel \citep{bagel}.
As shown in Fig.~\ref{fig:teaser} (3), both leading open- and closed-source models exhibit strong performance on prompts involving style and world knowledge, yet consistently struggle with logical reasoning that requires causal, contrastive, or other complex relational understanding.
Furthermore, open-source models show larger performance fluctuations across dimensions, particularly underperforming in the \textit{grammar} and \textit{action} dimensions. This highlights the {models'} difficulty in handling grammar-conditioned instructions and depicting dynamic or behavior-centric content accurately.
{To examine these weaknesses in greater detail, we conduct testpoint-level diagnostics and evaluate leading models on the challenging subset \ourbench-Hard, which increases score separation. Besides these diagnostic analyses, human validation shows close agreement between automatic and human-based rankings, and experiments with alternative models for benchmark construction and evaluation support the robustness of model comparisons.}
% !TEX root = ../main.tex
\begin{figure}[ht]
\centering
\includegraphics[width=0.95\textwidth]{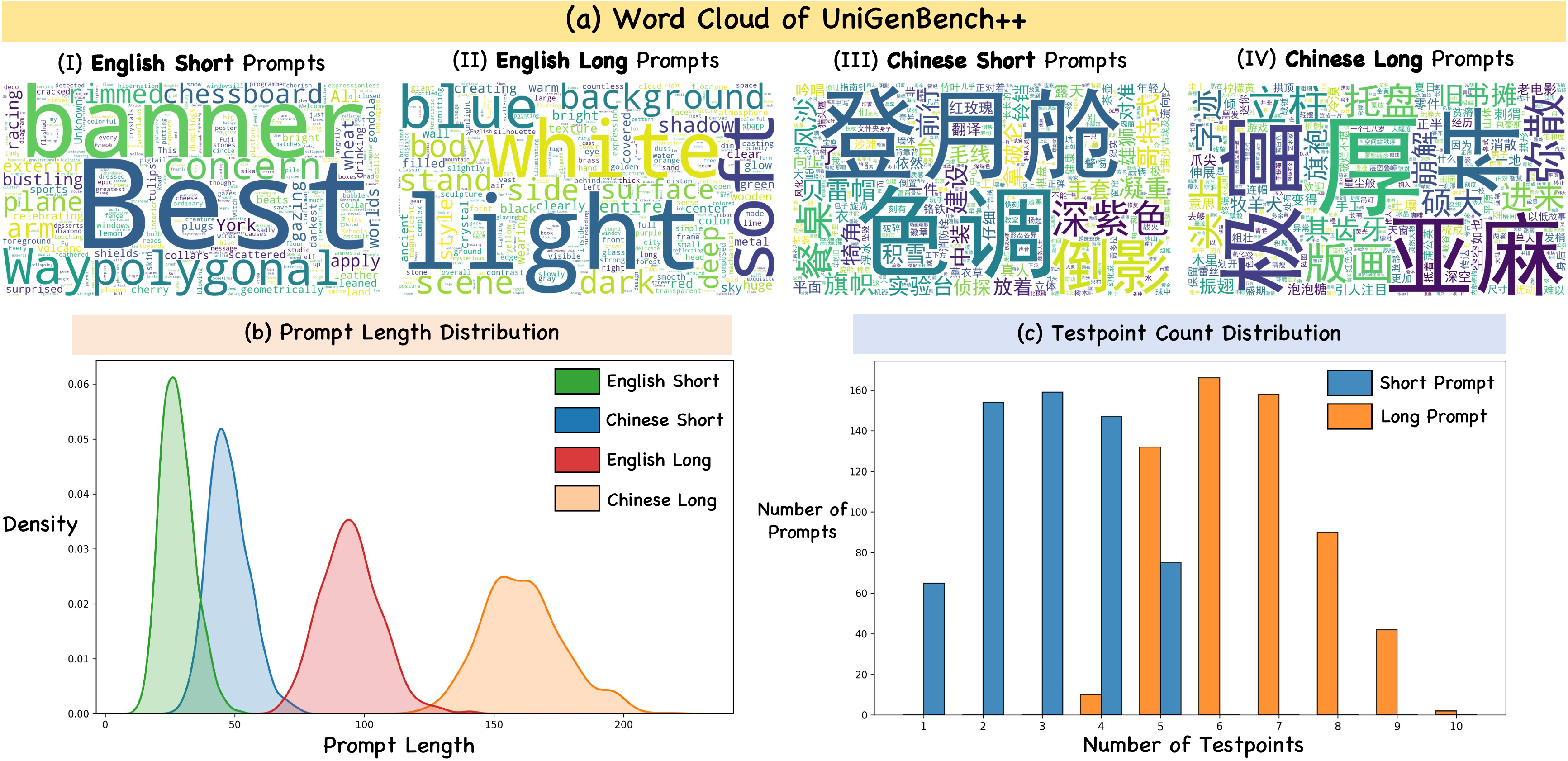}
% \vspace{-0.6cm}

\caption{\textbf{Benchmark Statistics.} (a) Word clouds for English and Chinese prompts in both short and long forms; (b) overall prompt length distribution; and (c) distribution of testpoint counts per prompt for short versus long versions.}
\label{fig:benchmark_info}
% \vspace{-0.4cm}
\end{figure}

The contributions of this paper are summarized as follows:
\begin{itemize}
    \item We propose \ourbench, a unified benchmark for text-to-image (T2I) semantic generation evaluation, covering comprehensive evaluation dimensions, diverse prompt themes, and rich subject categories. Each prompt is provided in both English and Chinese, and in short and long forms, assessing multiple test points, ensuring both coverage and efficiency.
    
    \item {We design a \textbf{streamlined, point-wise evaluation pipeline} for efficient and interpretable assessment of individual semantic requirements. Human validation and experiments with alternative benchmark-construction models and evaluators support the reliability of its judgments and the robustness of the resulting model comparisons.}

    \item {We provide a \textbf{dedicated offline evaluation model} that supports local assessment of T2I outputs for community use. Our evaluator improves agreement with human judgments over the untuned baseline and produces model rankings closely aligned with human assessment.}
    \item {We conduct \textbf{extensive bilingual and length-varied benchmarking} across open- and closed-source T2I models. Fine-grained diagnostics reveal distinct capability profiles and shared weaknesses in semantic generation, while the challenging subset \ourbench-Hard increases score separation among leading models.}
\end{itemize}
With its taxonomy-guided design, \ourbench{} is intended to provide a structured diagnosis of semantic generation capabilities across application-relevant scenarios. Its LLM-generated prompts enable systematic coverage of fine-grained semantic requirements, while their distribution may differ from that of real-world user requests. We hope that our benchmark could advance the development and evaluation of T2I models, driving further
improvements in semantic consistency across diverse fine-grained tasks and fostering deeper insights into
model performance across real-world scenarios.

% !TEX root = main.tex
\section{Related Work}
\noindent\textbf{Text-to-Image Generation.} 
Recent progress in text-to-image (T2I) generation is largely driven by two paradigms: diffusion-based and autoregressive (AR) models. Diffusion models dominate current practice due to their scalability and photorealistic synthesis, progressively denoising Gaussian noise conditioned on text, evolved from early GLIDE \citep{nichol2021glide} and Imagen \citep{imagen4} to powerful variants like Stable Diffusion \citep{sd}, FLUX \citep{flux}, and HiDream \citep{cai2025hidream}. In contrast, AR models generate images token by token via VQ-VAE \citep{van2017neural} compression and transformer decoding, as seen in the original DALL-E \citep{ramesh2021dalle} and CogView \citep{ding2021cogview}. Recent advances \citep{li2025onecat,team2024chameleon} enhance AR models with unified multimodal reasoning, while hybrid architectures like Bagel \citep{bagel} integrate both diffusion and AR to enable explicit reasoning before image generation.
With such rapid advances, evaluating T2I models, especially their semantic generation capability, has become a central challenge.

\noindent\textbf{Text-to-Image Benchmarks.}  
Prior studies commonly assess T2I models through compositional generation tests. For example, GenEval \citep{ghosh2023geneval} leverages object detection to rigorously verify whether generated images accurately reflect the spatial arrangements, numerical counts, and color attributes specified in the textual prompts. T2I-CompBench \citep{t2i-compbench} encompasses four core compositional categories and further extends these evaluations with detection-based metrics for spatial reasoning and numerical consistency. Several studies evaluate T2I models through specific knowledge domains, such as physical reasoning \citep{t2i-reasonbench} and general commonsense understanding \citep{niu2025wise,luo2025mmmg}.
However, the prompts used in these benchmarks are predominantly short and highly repetitive, which constrains semantic richness and expressiveness. Therefore, DPG-Bench \citep{dpg-bench} centers on assessing {models'} capability in dense prompts. TIIF-Bench \citep{wei2025tiif} offers both short and long variants of each prompt while preserving identical core semantics.

Despite their effectiveness, these benchmarks still suffer from coarse evaluation across limited dimensions and provide insufficient sub-dimension coverage. Moreover, the lack of diverse prompt scenarios and multilingual support further limits their ability to assess models in real-world application settings. To this end, we introduce \ourbench, a unified semantic-generation benchmark designed for fine-grained and comprehensive evaluation of T2I models.
% , featuring broad evaluation dimensions, diverse prompt scenarios, and rich subject categories.

% !TEX root = main.tex
\section{Benchmark}

\subsection{Overview}
With the rapid advancement of text-to-image (T2I) models, existing evaluation frameworks \citep{ghosh2023geneval,t2i-compbench,wei2025tiif,dpg-bench} have become increasingly insufficient. To be precise, (1) as summarized in Tab.~\ref{tab:bench_comparison}, they often overlook diversity in prompt scenarios and lack multilingual coverage, both of which are indispensable for real-world applicability. Consequently, their evaluations fall short in capturing a {model's} true applicability across diverse and contextually complex input conditions; (2) although existing benchmarks effectively assess a few broad dimensions, they still overlook several critical semantic aspects and lack systematic coverage and evaluation at the sub-dimension level, ultimately limiting their fine-grained diagnostic capability.

To this end, we propose \ourbench, a unified semantic evaluation benchmark for T2I generation. As summarized in Fig.~\ref{fig:teaser} and Tab.~\ref{tab:bench_comparison}, our benchmark offers several key advantages over existing studies:
\begin{itemize}
    \item \textbf{Rich prompt theme design.} Prompts are hierarchically organized into 5 primary themes and 20 sub-themes, spanning both real-world use cases and open-ended imaginative scenarios (Sec.~\ref{sec:themes}).
    \item \textbf{Comprehensive semantic dimension coverage.} It evaluates 10 primary dimensions and 27 sub-dimensions, enabling systematic diagnosis of diverse model capabilities. Despite its breadth, it requires only 600 prompts, each targeting {1--10} explicit test points, achieving a favorable balance between coverage and efficiency (Sec.~\ref{sec:dimensions}).
    \item \textbf{Bilingual and length-variant prompt and streamlined model evaluation.} All prompts are provided in both English and Chinese, each available in both short and long forms (Sec.~\ref{sec:prompt_construction}). Leveraging the world knowledge and fine-grained image understanding capabilities of Multi-modal Large Language Models (MLLMs), \ie Gemini-2.5-Pro, we design a fully streamlined pipeline for accurate and efficient model evaluation (Sec.~\ref{sec:model_evaluation}). 
    \item \textbf{Reliable evaluation model for offline assessment.} To facilitate community use, we train a robust evaluation model that supports offline assessment of T2I model outputs (Sec.~\ref{sec:offline_eval_model}).

    % Each prompt is composed according to predefined themes and test points, and every test point is explicitly paired with its in-context explanation . This design enables the evaluator model to perform deterministic, point-wise scoring, rather than relying on ambiguous multi-turn conversational reasoning .
    % \item \textbf{Bilingual and length-variant prompt support.} All prompts are provided in both English and Chinese, each available in both short and long forms, enabling systematic analysis of language sensitivity and length robustness.
    
\end{itemize}

\subsection{Prompt Themes and Subject Categories} \label{sec:themes}
We organize prompt themes around practical generation scenarios to broaden the coverage of application-relevant semantic requirements. The resulting taxonomy guides structured capability assessment across diverse scenarios.
Specifically, as illustrated in Fig.~\ref{fig:teaser} (1.a), we structure them into 5 primary categories and 20 finer sub-categories to ensure both breadth and practical relevance:
\begin{itemize}
    \item \textbf{Creative Divergence} covers open-ended imaginative ideation and broader forms of other abstract conceptual composition.
    \item \textbf{Art} encompasses a wide range of visual expression styles, including graphic renderings, photography-inspired depictions, sculptural aesthetics, and other fine-art formats.
    \item  \textbf{Illustration} is divided into copywriting-oriented visualization (\eg slogans or metaphors) and content-centric narrative illustration.
    \item \textbf{Film \& Story} accounts for settings across cinematic realism, speculative or science-fiction narratives, and animation-style storytelling.
    \item \textbf{Design} spans professional and commercial use cases such as advertising and e-commerce graphics, spatial layouts, game and UI prototyping, poster composition, IP and logo/icon creation, fashion concept design, and general-purpose design resource generation.
\end{itemize}

To facilitate understanding of each theme, we present representative prompts in Tab.~\ref{tab:prompt_theme_example}.

Based on a wide range of prompt themes, we further define a diverse set of subject categories to cover different types of entities. As illustrated in Fig.~\ref{fig:teaser} (1.d), these categories include \emph{animals}, \emph{objects}, \emph{anthropomorphic characters}, \emph{scenes}, as well as an \emph{Other} category for special or atypical entities (\eg robots appearing in science-fiction prompts). To this end, the benchmark can probe model capabilities on both common and unusual entities, providing insights into model strengths and weaknesses across diverse semantic scenarios.

The distribution of prompt themes and subject categories is illustrated in Fig.~\ref{fig:teaser} (1.a) and (1.d), respectively.

% !TEX root = ../main.tex
\begin{figure}[!t]
\centering
\includegraphics[width=0.9\textwidth]{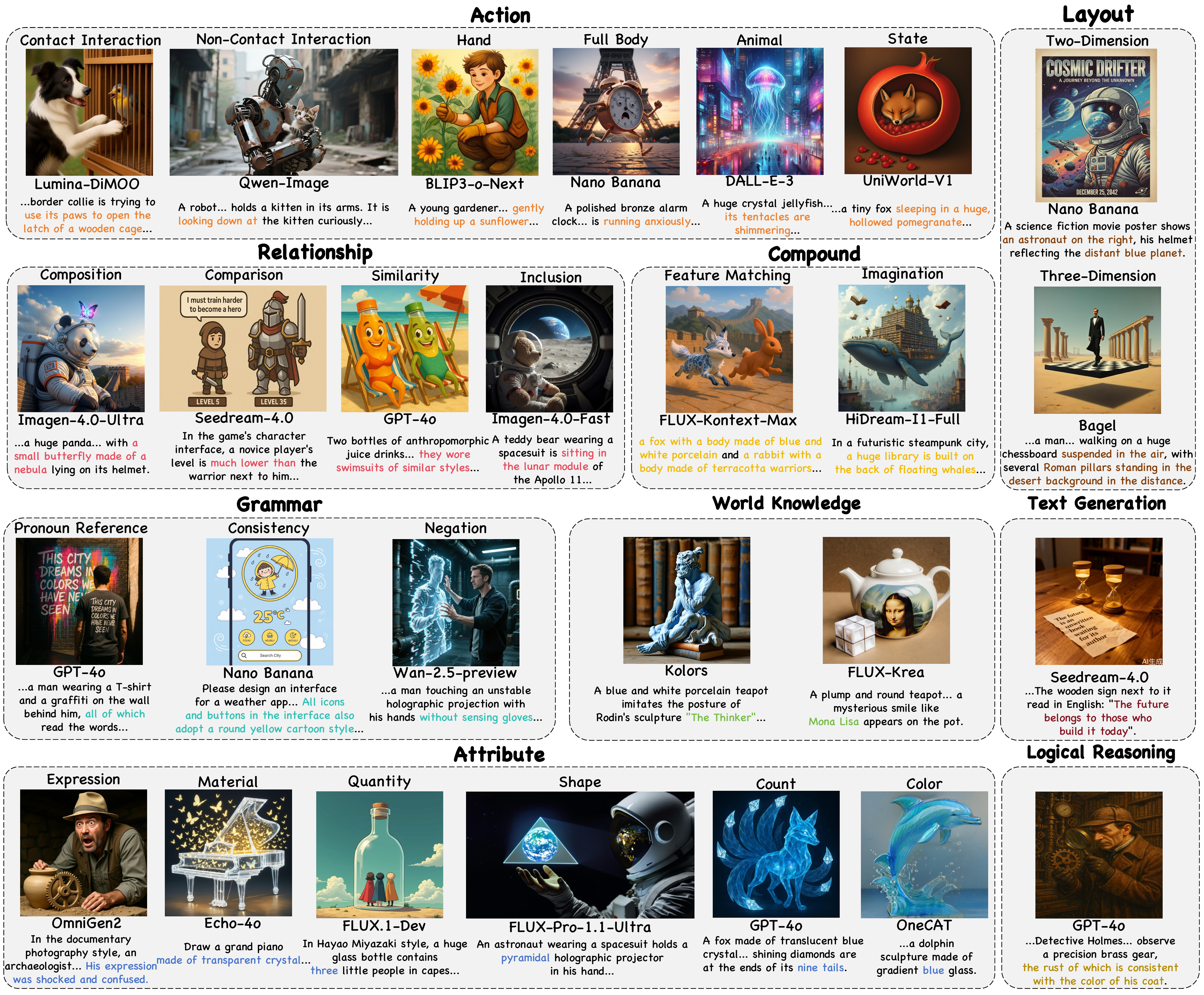}
% \vspace{-0.5cm}

\caption{\textbf{Qualitative Results of Evaluation Dimensions.} We present qualitative examples of T2I models evaluated across our specified dimensions.}
\label{fig:show_case}
% \vspace{-0.4cm}
% \vspace{-0.5cm}

\end{figure}

\subsection{Evaluation Dimensions}
\label{sec:dimensions}

Existing T2I models have demonstrated strong performance on several primary evaluation dimensions in current benchmarks. However, this surface-level success often masks their underlying weaknesses at the sub-dimension level, as coarse-grained metrics are insufficient to reveal fine-grained limitations in specific sub-aspects.
To address this gap, we decompose each major dimension into explicit and comprehensive sub-dimension-level test points. 
% This finer granularity enables a more diagnostic evaluation of model capabilities. 
Specifically, our benchmark organizes evaluation dimensions into 10 major categories, most of which encompass multiple subcategories:

\textbf{1. Style} evaluates the model's ability to generate images with coherent style and artistic expression. It considers both overall visual style and artistic genre, ensuring that the generated images exhibit plausible and consistent artistic characteristics.

\textbf{2. World Knowledge} examines the model's grasp of real-world concepts. It evaluates whether the model can generate content consistent with physical laws, cultural norms, geographical facts, and historical context.

\textbf{3. Attribute} assesses the model's understanding of object and scene characteristics, including:
\begin{itemize}
    \item \textbf{Quantity}: The number of objects or elements in a scene.
    \item \textbf{Expression}: Emotional states or facial expressions of humans or animals.
    \item \textbf{Material}: Surface properties of objects, such as wood, metal, or glass.
    \item \textbf{Color}: Accuracy and appropriateness of colors and color combinations.
    \item \textbf{Shape}: Geometric form and contour of objects.
    \item \textbf{Size}: Relative dimensions of objects within the scene.
\end{itemize}

\textbf{4. Compound} evaluates the model's ability to combine multiple concepts or features:
\begin{itemize}
    \item \textbf{Imagination}: Creativity in generating novel or non-realistic combinations.
    \item \textbf{Feature Matching}: Coherent integration of different elements and their attributes.
\end{itemize}

\textbf{5. Action} focuses on the dynamic behaviors and interactions of characters, animals, or objects:
\begin{itemize}
    \item \textbf{Contact Interaction}: Physical interactions between objects, such as touching and holding.
    \item \textbf{Non-contact Interaction}: Non-physical interactions like gazing.
    \item \textbf{Hand Actions}: Representation of hand gestures or manipulations.
    \item \textbf{Full-body Actions}: Depiction of whole-body movements of characters.
    \item \textbf{State}: Status or posture of objects or characters, such as sleeping, suspending, or running.
    \item \textbf{Animal Actions}: Behaviors specific to animals.
\end{itemize}

\textbf{6. Entity Layout} evaluates spatial arrangement and composition:
\begin{itemize}
    \item \textbf{Two-Dimensional Space}: Layout and relative positions of objects on a plane.
    \item \textbf{Three-Dimensional Space}:  Layout and relative positions of objects in three-dimensional space.
    
\end{itemize}

\textbf{7. Relationship} assesses the semantic and logical connections between objects:
\begin{itemize}
    \item \textbf{Composition}: Integration of multiple elements into a coherent whole.
    \item \textbf{Similarity}: Similarity in shape, color, or material between objects.
    \item \textbf{Comparison}: Differences and contrasts between objects.
    \item \textbf{Inclusion}: Containment or hierarchical relationships among objects.
\end{itemize}

\textbf{8. Logical Reasoning} measures the model's ability to reason about events, object attributes, understand causality, and contrastive expressions.

\textbf{9. Grammar} evaluates the model's understanding of textual and language-related expressions:
\begin{itemize}
    \item \textbf{Pronoun Reference}: Correct association between pronouns and their referents in the image.
    \item \textbf{Consistency}: Maintenance of coherent attributes, properties, or features across objects as described in the prompt.
    \item \textbf{Negation}: Accurate reflection of negation or exclusion expressions in the generated content.
\end{itemize}

\textbf{10. Text Generation} evaluates the model's ability to generate text content that is accurate, readable, and aligned with the requirements of the input prompt.

We provide qualitative examples of our evaluation dimensions in Fig.~\ref{fig:show_case}. Notably, in our benchmark, the distribution of test points differs between short and long prompts. Specifically, long prompts tend to have more attribute-related test points, as they provide more detailed and diverse descriptions of subjects, attributes, and scenes. The test point distribution for both is shown in Fig.~\ref{fig:teaser} (1.b) and (1.c).

% This hierarchical dimension system enables a detailed, multi-faceted evaluation of T2I models, covering artistic, factual, spatial, relational, logical, and textual aspects of image generation.

% !TEX root = ../main.tex
\begin{figure}[!t]
\centering
\includegraphics[height=0.65\textheight,keepaspectratio]{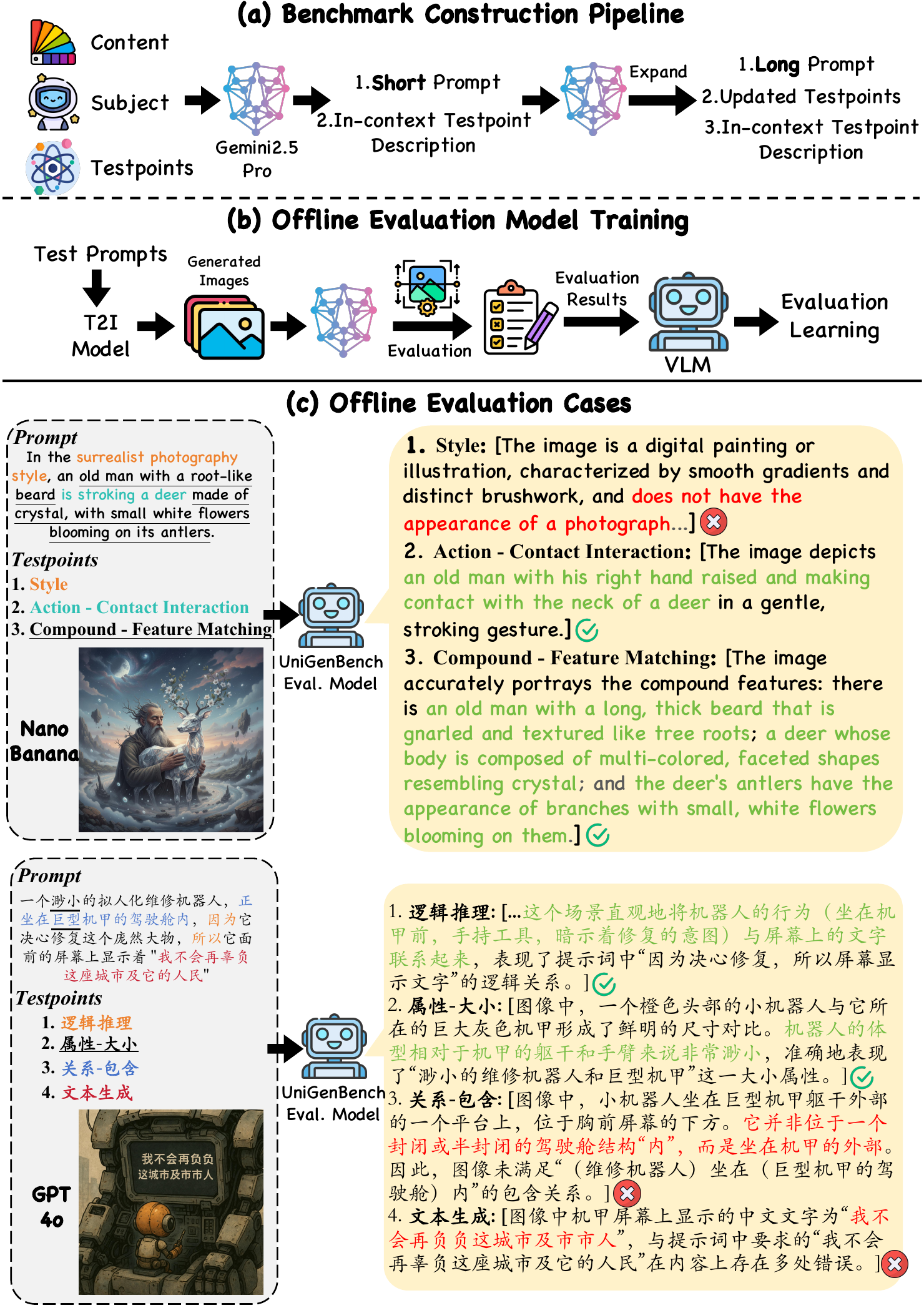}
\vspace{-0.3em}
\caption{\textbf{Pipeline of Benchmark Construction and Offline Evaluation Model Training.} (a) Benchmark construction pipeline; (b) Offline evaluation model training; (c) Offline evaluation cases.}
\label{fig:pipeline}
\vspace{-0.4em}
\end{figure}

\subsection{Bilingual and Length-variant Prompt Construction} \label{sec:prompt_construction}

\textbf{Bilingual Short Prompt Generation.} Let $\mathcal{T}$ denote the set of prompt \textit{themes}, $\mathcal{S}$ the set of \textit{subject categories}, and $\mathcal{C}$ the set of \textit{evaluation dimensions}. For each prompt construction step, a theme $t \sim \mathcal{T}$ and a subject category $s \sim \mathcal{S}$ are first sampled uniformly at random. Subsequently, a subset of $k$ test points ${c_1, \dots, c_k} \subset \mathcal{C}$, where $k \in [1,5]$, is selected to specify the targeted fine-grained test points.

Given the input tuple $(t, s, {c_1, \dots, c_k})$, the MLLM produces two outputs: (i) a pair of natural language prompts $(p^{\text{en}}, p^{\text{zh}})$ in English and Chinese, both adhering to the semantic constraints imposed by the selected theme $t$ and subject category $s$; and (ii) a structured description set ${d_1, \dots, d_k}$, where each element explicitly explains how the corresponding testpoint $c_i$ is instantiated within the generated prompts. Formally:
\begin{equation}
\big(p^{\text{en}}, p^{\text{zh}}, \{d_i\}_{i=1}^{k}\big)
\sim \mathrm{MLLM}_{\mathrm{gen}}\big(t, s, \{c_i\}_{i=1}^{k}\big),
\end{equation}

\textbf{Expanded to Long Prompt.}
To enrich the descriptive diversity and specificity of the generated prompts, we further expand each short prompt into a long-form prompt through a rewriting strategy. Given a short prompt $p^{\text{en}}$ or $p^{\text{zh}}$, we instruct the MLLM to generate an expanded version $\tilde{p}$ that satisfies two constraints: (i) the prompt theme, core subjects and their key attributes must be preserved, and (ii) attribute, scene, and background details may be further elaborated to enhance specificity and imagination. Formally,
\begin{equation}
\tilde{p} \sim \mathrm{MLLM}_{\mathrm{expand}}\big(p \mid r\big),
\end{equation}
where $r$ denotes the rewriting constraint.

However, expanding a prompt may introduce new semantic elements that are not covered by the original evaluation dimensions, or render some of the initial test points no longer applicable. To maintain consistency between the expanded prompt and its associated test points, we perform a second refinement step. Given the expanded prompt $\tilde{p}$ and the original test points $\{c_1, \dots, c_k\}$ with their descriptions $\{d_1, \dots, d_k\}$, we instruct the MLLM to revise the testpoint set by: (i) removing those no longer grounded in $\tilde{p}$; (ii) adding newly emerged test points, with a maximum allowance of five additional entries; and (iii) updating the in-context descriptions for all retained or newly added test points to reflect the semantics of $\tilde{p}$. Formally, the alignment process is defined as
\begin{equation}
\begin{aligned}
\big\{(\hat{c}_i,\hat{d}_i)\big\}_{i=1}^{k'}
&\sim \mathrm{MLLM}_{\mathrm{align}}\big(\cdot \mid \tilde{p},\,\{(c_i,d_i)\}_{i=1}^{k} \big), \\
k' &\le k + 5,
\end{aligned}
\end{equation}
where $k'$ is determined dynamically by the updated semantic scope of $\tilde{p}$. The resulting tuple
\[
\big(\tilde{p},\,\{(\hat{c}_i,\hat{d}_i)\}_{i=1}^{k'}\big)
\]
constitutes a semantically coherent long prompt paired with aligned and fine-grained evaluation targets.

The word clouds of both English and Chinese prompts in short and long forms are visualized in Fig.~\ref{fig:benchmark_info} (a). We also present statistics on the length distribution of prompts in Fig.~\ref{fig:benchmark_info} (b), as well as the distribution of test point counts between short and long prompts in Fig.~\ref{fig:benchmark_info} (c).

% !TEX root = ../main.tex
\begin{figure}[H]
\centering
\includegraphics[width=1\textwidth]{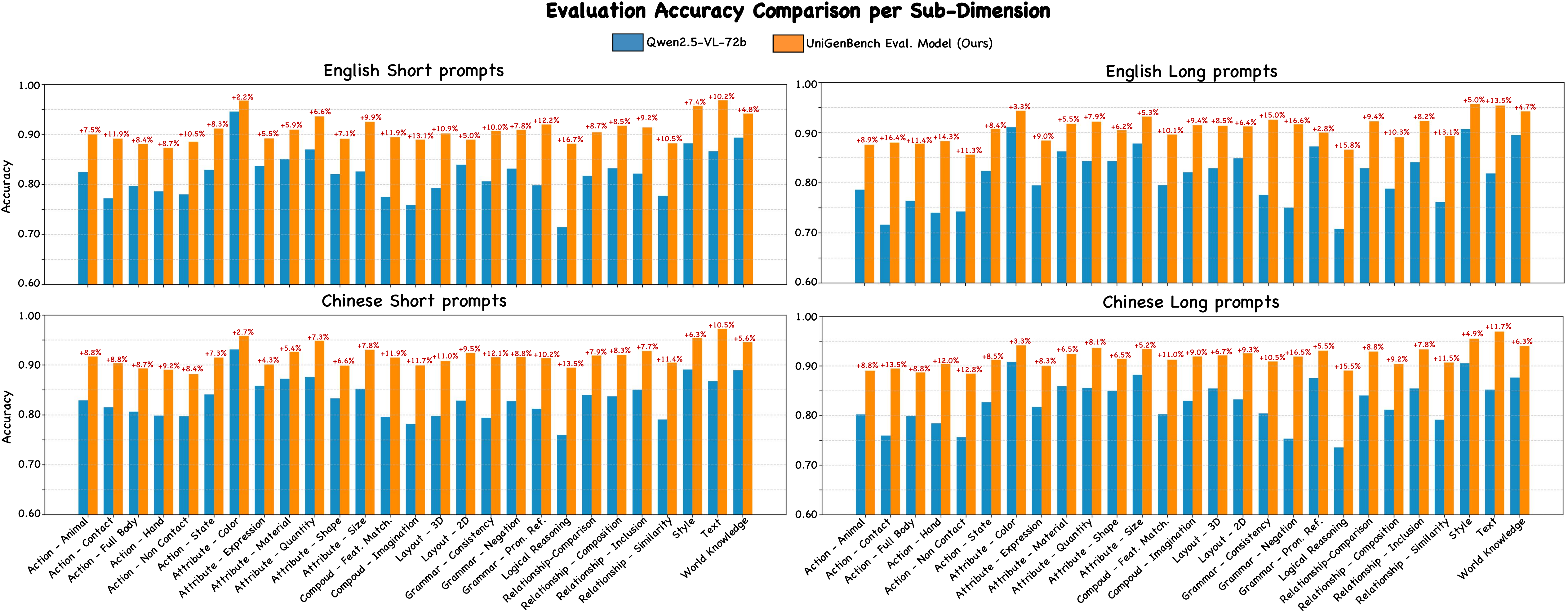}
% \vspace{-0.5cm}

\caption{\textbf{Evaluation Accuracy Comparison.} Our dedicated evaluation model demonstrates a significant improvement in evaluation accuracy across all test points compared to the commonly used offline evaluation VLM, \ie Qwen2.5-VL-72b.}
\label{fig:eval_acc}
% \vspace{-0.5cm}

\end{figure}

\subsection{T2I Model Evaluation}
\label{sec:model_evaluation}
To systematically evaluate the quality of model-generated images, we employ a MLLM, \ie Gemini-2.5-Pro, as an automatic evaluator. For each test prompt $p_i$, the corresponding generated image $x_i$ is paired with a set of fine-grained test points $\{c_{i,1}, \dots, c_{i,k}\}$ and their descriptions $\{d_{i,1}, \dots, d_{i,k}\}$. Since each test point corresponds uniquely to its description, we henceforth refer only to the descriptions $\{d_{i,j}\}$ for brevity. Then, the MLLM takes $(x_i, p_i, \{d_{i,j}\})$ as input and performs an independent assessment for each testpoint. For each $d_{i,j}$, it returns both a binary decision $r_{i,j} \in \{0,1\}$, indicating whether the requirement is satisfied, and a natural-language explanation $e_{i,j}$, which articulates the reasoning behind the judgment. This process is formally expressed as:
\begin{equation}
\big(\{r_{i,j}\}_{j=1}^{k}, \{e_{i,j}\}_{j=1}^{k}\big)
\sim \mathrm{MLLM}\big(\{r_{i,j}, e_{i,j}\}_{j=1}^{k} \mid x_i, p_i, \{d_{i,j}\}_{j=1}^{k}\big).
\end{equation}

Compared to scalar-only metrics, this formulation not only quantifies correctness but also reveals failure modes by exposing \emph{why} a testpoint is considered satisfied or violated. The availability of rationales $e_{i,j}$ further facilitates downstream error attribution. We provide an example evaluation case in Fig.~\ref{fig:teaser} (2).

Once all evaluation results are collected, we aggregate them at both the sub-dimension and primary-dimension levels. For each sub-dimension $c$, which groups semantically related test points, its score is defined as the ratio of satisfied instances to the total number of its occurrences across the benchmark:
\begin{equation}
R_c = \frac{\sum_{i,j} \mathbf{1}\{d_{i,j} \in c \text{ and } r_{i,j}=1\}}{\sum_{i,j} \mathbf{1}\{d_{i,j} \in c\}},
\end{equation}
where $\mathbf{1}\{\cdot\}$ denotes the indicator function. For the leaderboards, each primary dimension $C$ is scored by pooling its testpoint instances within each language--length setting. Let $n_c$ be the number of evaluated instances in sub-dimension $c$, $\mathcal{S}(C)$ the sub-dimensions in $C$, and $\mathcal{C}$ the set of 10 primary dimensions. The primary-dimension and setting scores are
\begin{equation}
R_C=\frac{\sum_{c\in\mathcal{S}(C)}n_c R_c}{\sum_{c\in\mathcal{S}(C)}n_c},
\qquad
R_{\mathrm{setting}}=\frac{1}{|\mathcal{C}|}\sum_{C\in\mathcal{C}}R_C.
\end{equation}

This hierarchical aggregation strategy enables multi-granular evaluation: it reflects fine-grained capability trends while also supporting concise reporting at a holistic level. Moreover, by separating binary correctness from explanatory evidence, our protocol provides both \emph{quantitative comparability} and \emph{qualitative interpretability}, which are crucial for diagnosing the strengths and weaknesses of T2I models at scale.

\subsection{Offline Evaluation Model Training}
\label{sec:offline_eval_model}

To facilitate convenient and cost-efficient evaluation for the community, we further train an \emph{offline evaluation model} that serves as a lightweight substitute for proprietary MLLMs during evaluation. Instead of querying a proprietary model online for every evaluation instance, our goal is to distill its scoring behavior into a compact model that can be executed locally without external API calls.

The supervision signals are constructed as described above from the online MLLM evaluator: for each image--prompt pair $(x_i,p_i)$ and testpoint description $\{d_{i,j}\}$, the reference outputs $(r_{i,j}, e_{i,j})$ produced by the MLLM are collected and assembled into target sequences for supervised fine-tuning.
Formally, given the tokenized target sequence $y_{i}$ associated with input $(x_i,p_i,\{d_{i,j}\})$, the training objective is:
\begin{equation}
\mathcal{L}(\theta)
= -\sum_{t=1}^{T_{i}} \log P_{\theta}\big(y_{i}^{(t)} \mid y_{i}^{(<t)},\, x_i, p_i, \{d_{i,j}\}_{j=1}^{k}\big),
\end{equation}
where $T_{i}$ is the length of $y_{i}$. This formulation allows the model to explicitly learn both binary judgment and explanatory reasoning through a language modeling objective.

At evaluation time, the offline evaluator can follow the same workflow as the original proprietary model-based assessment pipeline, producing decisions and explanatory rationales in a manner consistent with the online model.

% !TEX root = main.tex

\section{Experiment}

\subsection{Implementation Details}
\subsubsection{Benchmarking Models}

\textbf{Closed-source Models.} GPT-4o \citep{gpt-image-1}, Imagen-3.0/4.0-Ultra/Fast \citep{imagen4}, Nano Banana (Pro) \citep{nano-banana}, Seedream-3.0/4.0/4.5 \citep{seedream3,seedream4}, Wan2.2/2.5 \citep{wan-t2i}, Runway-Gen4 \citep{runway-gen4}, Recraft \citep{recraft}, DALL-E-3 \citep{dall-e}, FLUX-Pro-1.1-Ultra/Kontext-Max/Kontext-Pro \citep{flux}, HiDream-v2L \citep{hidream_v2L}, Stable-Image-Ultra \citep{stable_image_ultra}, FLUX-2-Pro/Flex/Max \citep{flux2}.

\noindent\textbf{Open-source Models.} Qwen-Image \citep{qwen-image}, Hunyuan-Image-2.1 \citep{hunyuan-image-2.1}, HiDream-I1-Full \citep{cai2025hidream}, Lumina-DiMOO \citep{lumina_dimoo}, Show-o2 \citep{show-o2}, Infinity \citep{han2025infinity}, OneCAT \citep{li2025onecat}, CogView4 \citep{cogview4}, X-Omni \citep{geng2025xomni}, MMaDA \citep{yang2025mmada}, FLUX.1-dev \citep{flux}, FLUX.1-Krea-dev \citep{flux-krea}, Echo-4o \citep{ye2025echo4o}, BLIP3-o (Next) \citep{blip3o}, UniWorld-V1 \citep{lin2025uniworld}, OmniGen2 \citep{omnigen2}, Bagel \citep{bagel}, Hunyuan-DiT \citep{li2024hunyuandit}, Janus series \citep{janus,ma2024janusflow,janus-pro}, Emu3 \citep{wang2024emu3}, Kolors \citep{kolors}, SDXL \citep{sdxl}, SD-3.5-Large \citep{sd35}, GLM-Image \citep{glm_image}, Z-Image (Turbo) \citep{z_image}, LongCat-Image \citep{longcat}, and FLUX.2-dev/Klein \citep{flux2}.

\subsubsection{Offline Evaluation Model} We adopt UnifiedReward-2.0-qwen-72b \citep{unifiedreward} as the base model and collect approximately 375K evaluation samples from Gemini-2.5-Pro. Of this, 300K is used for model training, and 75K is reserved for evaluation.
{
\subsection{Benchmarking Result Analysis}
\label{sec:benchmark_results}
% !TEX root = ../main.tex
{
\begin{table*}[t]
\centering
\scriptsize

\setlength{\tabcolsep}{4pt}
\renewcommand{\arraystretch}{0.98}
\caption{\textbf{Cross-setting summary of T2I model performance.} Only models evaluated in all four settings are included. The overall score is the equal-weight average of the four setting scores. Best and second-best scores are shown in \textbf{bold} and \underline{underlined}.}
\resizebox{\textwidth}{!}{%
\begin{tabular}{llccccc}
\toprule
\textbf{Model} & \textbf{Model availability} & \textbf{English short} & \textbf{English long} & \textbf{Chinese short} & \textbf{Chinese long} & \textbf{Overall} \\
\midrule
GPT-4o-1.5 & Closed & \textbf{95.77} & \textbf{95.41} & \textbf{95.62} & \textbf{96.12} & \textbf{95.73} \\
Nano Banana Pro & Closed & \underline{92.72} & \underline{94.20} & \underline{93.82} & \underline{95.42} & \underline{94.04} \\
GPT-4o & Closed & 92.48 & 92.63 & 91.02 & 90.51 & 91.66 \\
Seedream-4.5 & Closed & 89.70 & 91.38 & 89.58 & 93.12 & 90.94 \\
FLUX-2-Max & Closed & 90.85 & 92.18 & 88.14 & 89.80 & 90.24 \\
FLUX-2-Flex & Closed & 89.35 & 90.43 & 87.62 & 89.19 & 89.15 \\
Seedream-4.0 & Closed & 87.35 & 89.77 & 87.31 & 90.35 & 88.69 \\
FLUX-2-Pro & Closed & 88.35 & 90.10 & 85.40 & 87.11 & 87.74 \\
Imagen-4.0-Ultra & Closed & 91.65 & 90.95 & 83.08 & 83.86 & 87.39 \\
FLUX.2-dev & Open & 84.76 & 90.31 & 81.44 & 86.12 & 85.66 \\
Nano Banana & Closed & 87.29 & 88.82 & 80.45 & 83.17 & 84.93 \\
Z-Image & Open & 78.10 & 86.77 & 81.69 & 89.17 & 83.93 \\
Imagen-4.0 & Closed & 85.84 & 85.34 & 79.52 & 79.90 & 82.65 \\
Qwen-Image & Open & 78.36 & 83.94 & 81.04 & 86.91 & 82.56 \\
Seedream-3.0 & Closed & 78.41 & 80.99 & 81.68 & 86.14 & 81.80 \\
Wan2.5 & Closed & 77.87 & 84.56 & 78.86 & 84.36 & 81.41 \\
Hunyuan-Image-2.1 & Open & 74.64 & 82.19 & 77.76 & 87.01 & 80.40 \\
FLUX.2-Klein-base-9b & Open & 79.35 & 86.45 & 73.81 & 81.41 & 80.25 \\
FLUX.2-Klein-9b & Open & 78.28 & 85.06 & 75.19 & 81.74 & 80.07 \\
LongCat-Image & Open & 73.54 & 81.28 & 75.97 & 83.14 & 78.48 \\
Z-Image-Turbo & Open & 71.40 & 80.72 & 74.18 & 83.69 & 77.50 \\
FLUX-Kontext-Max & Closed & 80.00 & 80.88 & 71.85 & 75.24 & 76.99 \\
Echo-4o & Open & 69.12 & 76.41 & 72.40 & 78.31 & 74.06 \\
GLM-Image & Open & 67.23 & 75.48 & 70.57 & 79.11 & 73.10 \\
DALL-E-3 & Closed & 68.85 & 70.82 & 67.93 & 71.16 & 69.69 \\
Bagel & Open & 59.91 & 71.26 & 65.69 & 75.75 & 68.15 \\
Wan2.2-Plus & Closed & 64.82 & 68.76 & 66.96 & 70.05 & 67.65 \\
OmniGen2 & Open & 63.09 & 71.39 & 63.20 & 70.75 & 67.11 \\
Lumina-DiMOO & Open & 71.12 & 71.81 & 58.35 & 63.80 & 66.27 \\
HiDream-I1-Full & Open & 71.36 & 74.25 & 50.65 & 50.70 & 61.74 \\
CogView4 & Open & 56.00 & 67.68 & 55.14 & 68.09 & 61.73 \\
OneCAT & Open & 58.28 & 62.80 & 56.77 & 61.40 & 59.81 \\
BLIP3-o & Open & 59.57 & 61.01 & 59.25 & 59.25 & 59.77 \\
Recraft & Closed & 62.63 & 60.93 & 57.67 & 56.90 & 59.53 \\
X-Omni & Open & 53.77 & 67.00 & 53.69 & 62.18 & 59.16 \\
BLIP3-o-Next & Open & 65.15 & 71.03 & 44.48 & 54.55 & 58.80 \\
Kolors & Open & 46.07 & 53.60 & 58.80 & 65.12 & 55.90 \\
Janus-Pro & Open & 61.36 & 71.11 & 30.83 & 60.21 & 55.88 \\
Hunyuan-DiT & Open & 51.38 & 54.88 & 53.36 & 55.57 & 53.80 \\
Janus & Open & 51.60 & 60.37 & 30.98 & 33.63 & 44.14 \\
MMaDA & Open & 41.35 & 40.10 & 44.00 & 50.61 & 44.02 \\
UniWorld-V1 & Open & 63.11 & 69.60 & 15.21 & 21.50 & 42.35 \\
Emu3 & Open & 45.42 & 50.95 & 33.91 & 35.95 & 41.56 \\
Janus-Flow & Open & 47.10 & 54.80 & 20.93 & 23.09 & 36.48 \\
\bottomrule
\end{tabular}%
}
\label{tab:cross_setting_summary}
\end{table*}
}

Tab.~\ref{tab:cross_setting_summary} provides a compact comparison of the 44 models evaluated in all four language--length settings; the setting-specific primary- and fine-grained-dimension results are reported in the appendix. GPT-4o-1.5 ranks first in every setting and achieves the highest four-setting average (95.73), followed by Nano Banana Pro (94.04). GPT-4o, Seedream-4.5, and FLUX-2-Max form the next group, each averaging above 90. Among open-source models, FLUX.2-dev obtains the strongest four-setting average (85.66), followed by Z-Image (83.93) and Qwen-Image (82.56). The frontier therefore remains led by closed-source models, although the strongest open-source models overlap with the middle of the closed-source range.

The four setting scores also expose behavior hidden by a single average. Several models are comparatively stable across settings, whereas others show pronounced language sensitivity. For example, Imagen-4.0-Ultra decreases from 91.65/90.95 in English to 83.08/83.86 in Chinese, while Z-Image increases from 78.10/86.77 to 81.69/89.17. Large differences are not confined to weaker models: they occur among both open- and closed-source models and motivate the cross-language analysis in Sec.~\ref{sec:cross_setting_analysis}. Long prompts also frequently yield higher scores than short prompts, but this pattern should not be interpreted as a general causal benefit. On strictly matched testpoints (Tab.~\ref{tab:cross_setting_robustness}), the long--short difference is $+5.17$ points in English (95\% CI: $+2.03$ to $+8.64$), but only $+0.86$ points in Chinese (95\% CI: $-0.81$ to $+2.63$).

These aggregate results establish broad performance tiers but do not reveal how models vary across settings or which capabilities distinguish similarly ranked models. The following analyses therefore examine item difficulty, uncertainty, challenging items, cross-setting consistency, and performance profiles after controlling for general capability. Full primary-dimension leaderboards for the four settings are provided in Tabs.~\ref{tab:benchmark_en_short_overall}--\ref{tab:benchmark_zh_long_overall}, and the corresponding 27-dimension results are provided in Tabs.~\ref{tab:benchmark_en_short}--\ref{tab:benchmark_zh_long}.
}

{
\subsection{Ceiling Effects, Uncertainty, and Discriminative Power}
\label{sec:ceiling_analysis}
The cross-setting summary provides an overall comparison of T2I models, but it does not fully characterize how benchmark difficulty is distributed across individual testpoints or how reliably closely ranked models can be distinguished. We therefore complement the aggregate results with an item-level analysis of benchmark difficulty, dimension-wise variation, ranking uncertainty, and performance on a challenging subset. This analysis provides a more detailed view of where current models perform consistently and where meaningful differences remain.

\textbf{Analysis protocol.} We analyze all 1,937,567 binary testpoint judgments from 409,674 generated images in the generation-level evaluation records. To avoid treating repeated generations as independent benchmark items, an item is defined by its language--length split, prompt, and testpoint. For each T2I model, we average the binary pass/fail scores across repeated generations, yielding one item-level pass rate per model. Item-level analyses cover the 35 models with generation-level records available across all four language--length settings. The item-level cohort yields 11,480 unique items. We define the frontier group as the 10 highest-scoring models within this 35-model cohort and use the remaining 25 models as the reference group.

\begin{table}[t]
\centering
\scriptsize

\setlength{\tabcolsep}{4.0pt}
\renewcommand{\arraystretch}{0.94}
\caption{\textbf{Item-level pass-rate distribution.} Easy: pass rate $\geq0.90$; Perfect: pass rate $=1.00$; Hard: pass rate $\leq0.50$.}
\begingroup
\fontsize{11.04}{13}\selectfont
\begin{tabular*}{\columnwidth}{@{\extracolsep{\fill}}lrrrrr}
\toprule
\textbf{Item set / split} & \textbf{$N$} & \textbf{Mean} & \textbf{Easy (\%)} & \textbf{Perfect (\%)} & \textbf{Hard (\%)} \\
\midrule
All models     & 11,480 & 0.714 & 25.1 &  2.9 & 19.1 \\
Frontier group & 11,480 & 0.892 & 73.6 & 46.0 &  6.6 \\
\midrule
English--Short &  1,814 & 0.715 & 32.2 &  5.0 & 21.9 \\
English--Long  &  3,926 & 0.775 & 42.5 &  6.2 & 14.8 \\
Chinese--Short &  1,814 & 0.647 & 12.6 &  0.1 & 26.4 \\
Chinese--Long  &  3,926 & 0.685 & 10.1 & $<0.1$ & 18.8 \\
\bottomrule
\end{tabular*}
\endgroup
\label{tab:item_difficulty}
\end{table}

\textbf{Item-level difficulty distribution.} Tab.~\ref{tab:item_difficulty} summarizes the distribution of item-level pass rates across the benchmark. Across all models, the mean item pass rate is 0.714; 25.1\% of items have a pass rate of at least 0.90, whereas 19.1\% remain at or below 0.50. Only 2.9\% attain a pass rate of 1.00 across all evaluated models and repeated generations. For the frontier group, the distribution shifts toward higher pass rates: 73.6\% of items reach at least 0.90, and 46.0\% reach 1.00. High pass rates are thus more concentrated among frontier models, while the complete model set exhibits a broader range of item difficulty. The distribution also varies across settings: the proportion of easy items ranges from 10.1\% for Chinese long prompts to 42.5\% for English long prompts. The following dimension-level analysis further localizes the remaining variation among models.

\begin{table}[t]
\centering
\scriptsize

\setlength{\tabcolsep}{3.0pt}
\renewcommand{\arraystretch}{0.96}
\caption{\textbf{Between-model variation by dimension (\%).} Each entry reports mean/standard deviation (SD)/range across model-level dimension scores in the indicated group.}
\resizebox{\columnwidth}{!}{%
\begin{tabular}{lcc@{\hspace{7pt}}lcc}
\toprule
\textbf{Dimension} & \textbf{All models} & \textbf{Frontier group} & \textbf{Dimension} & \textbf{All models} & \textbf{Frontier group} \\
\cmidrule(lr){2-3}\cmidrule(lr){5-6}
& \multicolumn{2}{c}{\textbf{Mean/SD/range}} & & \multicolumn{2}{c}{\textbf{Mean/SD/range}} \\
\midrule
Style           & 88.7/10.8/51.3 & 97.7/1.4/4.4   & Entity Layout      & 71.9/17.6/59.1 & 90.1/3.6/12.2 \\
World Knowledge & 81.8/16.3/57.8 & 96.2/1.5/3.9   & Relationship       & 70.1/17.1/59.4 & 89.2/4.1/11.9 \\
Attribute       & 73.0/16.2/57.5 & 90.3/2.7/8.4   & Logical Reasoning  & 50.5/19.4/69.0 & 74.4/9.7/28.6 \\
Compound        & 63.6/19.5/63.7 & 86.3/5.8/15.9  & Grammar            & 67.4/13.2/53.6 & 81.8/8.1/23.5 \\
Action          & 65.0/18.0/61.6 & 86.0/4.5/16.0  & Text Generation    & 30.7/30.7/94.8 & 67.3/19.9/56.1 \\
\bottomrule
\end{tabular}}
\label{tab:dimension_discrimination}
\end{table}

\textbf{Between-model variation across dimensions.} Tab.~\ref{tab:dimension_discrimination} examines how strongly model performance differs within each evaluation dimension. For these dimension profiles, we equally average sub-dimension scores within each primary dimension and then average the four settings. For each dimension and model group, we compute the mean, standard deviation (SD), and range across the models' dimension scores; a larger SD or range therefore indicates greater separation among models on that capability. Within the frontier group, Style and World Knowledge show little between-model variation, with SDs of 1.4 and 1.5 percentage points, respectively. In contrast, Logical Reasoning, Grammar, and Text Generation exhibit SDs of 9.7, 8.1, and 19.9 points and ranges of 28.6, 23.5, and 56.1 points. Compound and Action also retain ranges of approximately 16 points. These dimension-level distributions complement the aggregate scores by identifying the capabilities that most clearly differentiate current models.

\begin{table}[t]
\centering
\scriptsize

\setlength{\tabcolsep}{2.4pt}
\renewcommand{\arraystretch}{0.96}
\caption{\textbf{Bootstrap uncertainty and performance on \ourbench-Hard.} Full reports the benchmark score with its 95\% bootstrap confidence interval (CI) and central 95\% rank interval among the 10 frontier models, using the 2.5th and 97.5th percentiles of ranks across 10,000 prompt-cluster resamples. Hard reports the \ourbench-Hard score, with $\Delta$ denoting Hard minus Full in percentage points.}
\resizebox{\columnwidth}{!}{%
\begin{tabular}{lcc@{\hspace{6pt}}lcc}
\toprule
\textbf{Model} & \textbf{Full [CI; rank]} & \textbf{Hard [$\Delta$]} & \textbf{Model} & \textbf{Full [CI; rank]} & \textbf{Hard [$\Delta$]} \\
\midrule
Nano Banana Pro  & 94.04 [93.59--94.49; 1--1] & 88.30 [$-5.74$] & Nano Banana      & 84.93 [84.29--85.57; 5--6] & 74.96 [$-9.97$] \\
GPT-4o           & 91.66 [91.11--92.20; 2--2] & 84.90 [$-6.76$] & Imagen-4.0       & 82.65 [81.96--83.33; 7--8] & 69.32 [$-13.33$] \\
Seedream-4.0     & 88.69 [88.02--89.36; 3--3] & 77.77 [$-10.93$] & Qwen-Image       & 82.56 [81.73--83.39; 7--8] & 66.38 [$-16.18$] \\
Imagen-4.0-Ultra & 87.39 [86.78--87.97; 4--4] & 78.36 [$-9.02$] & Seedream-3.0     & 81.80 [80.94--82.66; 8--10] & 65.90 [$-15.91$] \\
FLUX.2-dev       & 85.66 [84.91--86.42; 5--6] & 73.22 [$-12.44$] & Wan2.5           & 81.41 [80.64--82.20; 9--10] & 67.38 [$-14.03$] \\
\bottomrule
\end{tabular}}
\label{tab:bootstrap_hard}
\end{table}

\textbf{Uncertainty and ranking stability.} We perform 10,000 percentile-bootstrap resamples, stratified by the four language--length settings, with the prompt as the cluster. Each resample retains all testpoints and model outputs associated with a selected prompt, recomputes the benchmark score within each setting, and then equally averages the four setting scores before reranking models. As reported in Tab.~\ref{tab:bootstrap_hard}, the leading four models each have a single-rank central 95\% interval, indicating high ranking stability without implying that their ranks are invariant in every resample. FLUX.2-dev and Nano Banana have overlapping 95\% rank intervals of 5--6; Imagen-4.0 and Qwen-Image span ranks 7--8, while Seedream-3.0 and Wan2.5 span 8--10 and 9--10, respectively. These intervals summarize ranking uncertainty under prompt-level resampling within the 10-model frontier group.

\textbf{Performance on a challenging subset.} We further examine frontier models on a challenging subset selected without using the frontier models' item-level outcomes. Specifically, item difficulty is estimated using only the non-frontier reference models, and items with a mean reference-group pass rate $\leq0.60$ are retained. The resulting \ourbench-Hard subset contains 4,290 items (37.4\% of the benchmark), spanning all four language--length settings and all 10 primary dimensions. As shown in Tab.~\ref{tab:bootstrap_hard}, scores range from 65.90 to 88.30 on \ourbench-Hard, compared with 81.41 to 94.04 on the full benchmark. The frontier mean decreases from 86.08 to 74.65, while the between-model SD increases from 4.33 to 7.77 points. The rankings on the full benchmark and \ourbench-Hard remain strongly correlated ($\rho=0.939$), so the subset increases score separation while retaining a broadly consistent ordering.
}

{
\subsection{Cross-Setting Consistency and Diagnostic Analysis}
\label{sec:cross_setting_analysis}
Beyond aggregate performance, we examine how model behavior varies across languages and prompt lengths and how the benchmark's dimensions and testpoints contribute to its diagnostic structure. The cross-language analysis includes models evaluated in all four language--length settings. For each language, we equally average sub-dimension scores within primary dimensions and then average the primary dimensions, with equal weights for short- and long-prompt results. To isolate prompt-length effects on shared semantic requirements, we form short--long pairs only when the testpoints have the same language, prompt index, canonical dimension, and normalized description, with unchanged metadata and complete model coverage.

\begin{table}[t]
\centering
\scriptsize

\setlength{\tabcolsep}{3.0pt}
\renewcommand{\arraystretch}{0.96}
\caption{\textbf{Cross-setting consistency.} Top: agreement between the English and Chinese hierarchical rankings. Bottom: long-minus-short pass-rate changes on strictly matched testpoints, in percentage points (pp), with 95\% CIs from 10,000 prompt-cluster bootstrap resamples.}
\resizebox{\columnwidth}{!}{%
\begin{tabular}{lccccc}
\toprule
\textbf{Analysis} & \textbf{$N$} & \textbf{Spearman $\rho$} & \textbf{Kendall $\tau$} & \textbf{Top-3/5 overlap} & \textbf{Max. rank shift} \\
\midrule
English vs. Chinese rank & 35 models & 0.751 & 0.573 & 2/3; 3/5 & 20 \\
\midrule
\textbf{Length comparison} & \textbf{Matched testpoints} & \textbf{Prompt pairs} & \textbf{Mean change (pp)} & \textbf{95\% CI} & \textbf{Improved/tied/declined (\%)} \\
English long--short & 67  & 61  & $+5.17$ & $[+2.03,+8.64]$ & 18.9/71.6/9.5 \\
Chinese long--short & 126 & 107 & $+0.86$ & $[-0.81,+2.63]$ & 16.1/70.6/13.3 \\
\bottomrule
\end{tabular}}
\label{tab:cross_setting_robustness}
\end{table}

\textbf{Cross-language consistency.} Tab.~\ref{tab:cross_setting_robustness} shows a moderately strong correspondence between the English and Chinese rankings ($\rho=0.751$, $\tau=0.573$). Two models are shared by the top-three sets and three by the top-five sets, and the two highest-ranked models retain their positions in both languages. At the same time, the maximum shift reaches 20 positions, indicating substantial language-specific variation for some models. The bilingual settings therefore reveal both a common capability ordering and differences in how individual models handle English and Chinese prompts.

\textbf{Prompt-length comparison.} On semantically matched testpoints, the effect of longer prompts differs between the two languages. In English, the average pass rate is 5.17 percentage points higher for long prompts (95\% CI: $+2.03$ to $+8.64$). The corresponding change in Chinese is $+0.86$ points, with a confidence interval spanning zero (95\% CI: $-0.81$ to $+2.63$). Ties account for most model--testpoint comparisons because each testpoint score averages four binary generation-level judgments; among the remaining comparisons, improvements are more frequent than declines in both languages. These results characterize prompt length as a language-dependent factor rather than a uniformly positive or negative source of performance change.

\begin{table}[t]
\centering
\scriptsize

\setlength{\tabcolsep}{4.0pt}
\renewcommand{\arraystretch}{0.96}
\caption{\textbf{Length-binned analysis.} All evaluated prompts are binned into quintiles by token length. Per bin, we report the mean testpoint pass rate over all evaluated models and the discrimination between model groups, measured as the mean score gap between the top and bottom 27\% of models ranked by hierarchical score. Gaps are in percentage points (pp).}
\begin{tabular*}{\columnwidth}{@{\extracolsep{\fill}}lccc@{}}
\toprule
\textbf{Length quintile} & \textbf{Mean tokens} & \textbf{Mean pass rate (\%)} & \textbf{Top--bottom gap (pp)} \\
\midrule
Q1 (shortest) & 14  & 78.2 & 21.5 \\
Q2            & 28  & 74.6 & 22.8 \\
Q3            & 47  & 70.1 & 23.4 \\
Q4            & 82  & 65.3 & 22.1 \\
Q5 (longest)  & 131 & 61.8 & 20.7 \\
\bottomrule
\end{tabular*}
\label{tab:length_binned_analysis}
\end{table}

\textbf{Length-binned analysis.} We group all evaluated prompts into token-length quintiles and compare mean testpoint pass rates and score gaps between the top and bottom 27\% of models (Tab.~\ref{tab:length_binned_analysis}). Pass rates decrease from 78.2\% to 61.8\%, while model-group gaps remain substantial at 20.7--23.4 percentage points. Unlike the matched-testpoint comparison, this analysis describes variation across the full prompt set rather than length effects under fixed semantic requirements. It shows increasing difficulty with prompt length alongside sustained model separation across bins.

\begin{figure*}[t]
\centering
\includegraphics[width=0.90\textwidth]{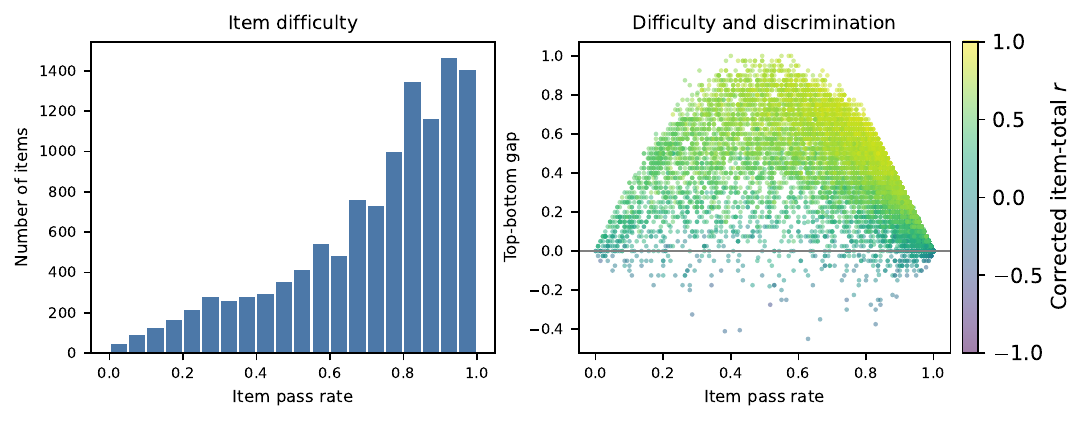}
\caption{\textbf{Testpoint difficulty and discrimination in \ourbench.} The left panel shows the testpoint pass-rate distribution; the right panel relates testpoint difficulty to the top--bottom discrimination gap, with color denoting corrected testpoint--total correlation.}
\label{fig:w3_item_diagnostics}
\end{figure*}

\begin{figure*}[t]
\centering
\includegraphics[width=0.9\textwidth]{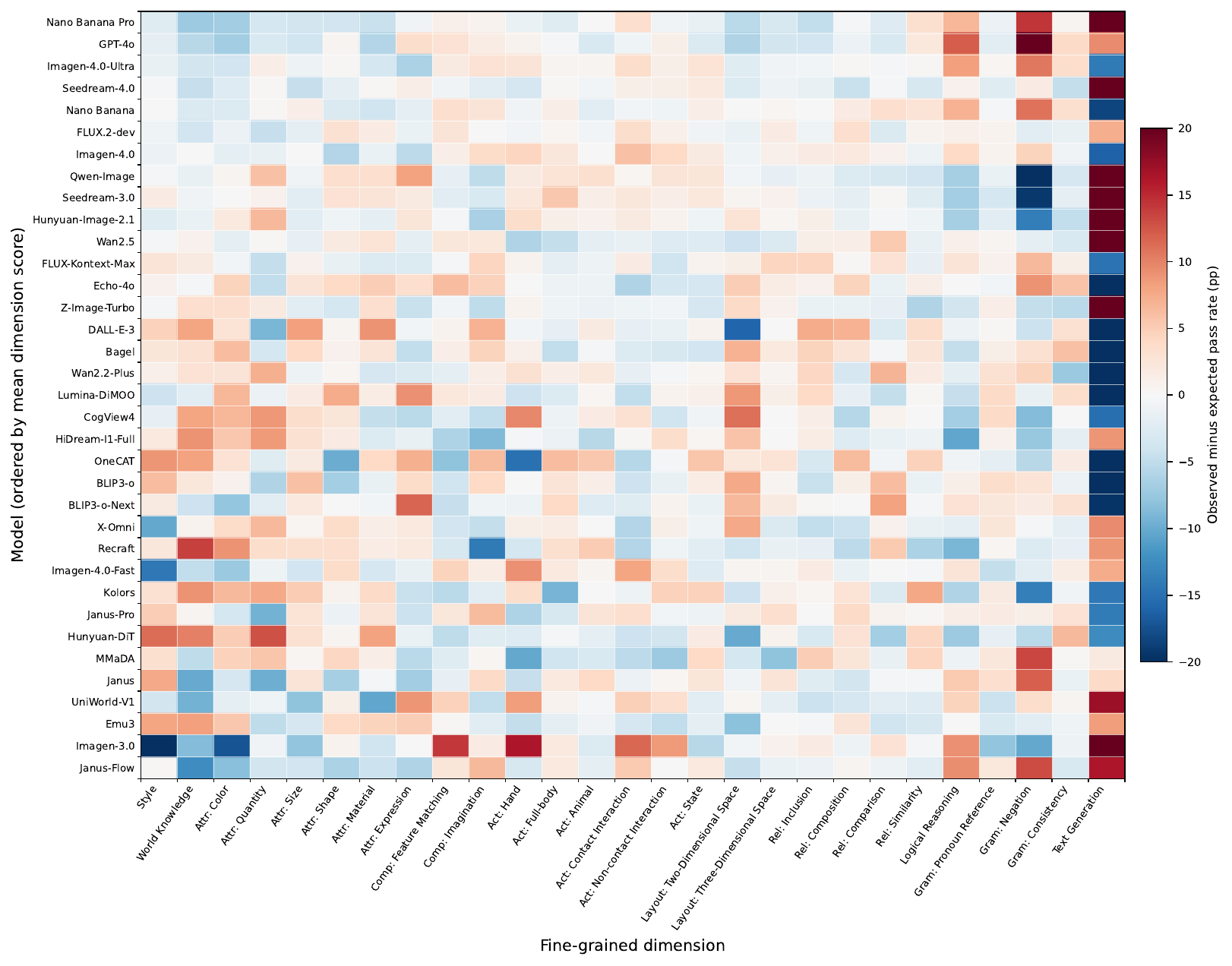}
\caption{\textbf{Residual capability profiles.} Cells show observed minus predicted dimension scores in percentage points (pp), with predictions based on mean performance on the other 26 dimensions. Positive and negative values indicate relative strengths and weaknesses, respectively.}
\label{fig:residual_profiles}
\end{figure*}

\textbf{Shared performance and dimension-specific profiles.} With dimension scores averaged equally across the four settings, we find that models performing well in one dimension tend to perform well in others (mean Spearman $\rho=0.906$ across 351 dimension pairs; median: 0.945). To quantify this shared trend, we regress each dimension score on the mean of the other 26 dimensions, explaining 89.6\% of cross-model variance on average. We then construct residual capability profiles from the differences between observed and fitted scores to examine where individual models depart from this trend. Appendix~\ref{sec:residual_profile_analysis} details the regression and residual-profile calculations.

Fig.~\ref{fig:residual_profiles} shows that models with similar mean performance can nevertheless have distinct residual profiles. For example, FLUX.2-dev and Nano Banana differ by only 0.32 percentage points in mean dimension score, yet their profiles have a root-mean-square distance of 6.29 points. Such differences are particularly pronounced in Text Generation, which has the largest residual standard deviation across models (20.17 points; $R^2=0.567$). Negation and Style also show lower explained variance ($R^2=0.387$ and 0.621), highlighting variation in text rendering, negation, and style that is less apparent from aggregate rankings.

\textbf{Testpoint difficulty and discrimination.} We measure each testpoint's discrimination as the pass-rate gap between the top and bottom 27\% of models ranked by hierarchical score across the four settings. The mean gap is 37.5 percentage points, and the mean corrected testpoint--total correlation, excluding the focal testpoint from the same-setting total, is 0.579. Fig.~\ref{fig:w3_item_diagnostics} shows that intermediate-difficulty testpoints generally provide the greatest separation, whereas near-universally passed testpoints are less discriminative. Together with Tab.~\ref{tab:item_difficulty}, these results characterize a broad difficulty range with substantial testpoint-level discrimination.
}

\subsection{Offline Evaluation Model}
Existing benchmarks \citep{niu2025wise,wei2025tiif} typically use Vision-Language Models (VLMs) like Qwen2.5-VL-72b \citep{bai2025qwen2} for offline generalization evaluation. However, compared to closed-source models, the agreement of these models with the online evaluator often falls short. Specifically, in our benchmark, we observed that Qwen2.5-VL-72b performs reasonably well on relatively simple dimensions such as attribute-color and facial expressions. However, its performance becomes unreliable on more complex dimensions like grammar-consistency and action-contact. To address this, we train a dedicated evaluation model, and its teacher-fidelity results, compared to Qwen2.5-VL-72b, are shown in Fig.~\ref{fig:eval_acc}. As demonstrated, our model more faithfully reproduces Gemini-2.5-Pro judgments across short and long Chinese and English prompts. Both English and Chinese qualitative evaluation cases are provided in Fig.~\ref{fig:pipeline} (c). {We further assess both the online and offline evaluators against independent human judgments in Sec.~\ref{sec:human_validation}.}

{
\subsection{Human Validation of the \ourbench\ Evaluators}
\label{sec:human_validation}
\textbf{Study design.} We compare Gemini-2.5-Pro~\citep{gemini2.5pro} with human judgments on 1,200 prompt--image--testpoint triplets, sampling three per combination of language (English/Chinese), prompt length (short/long), 10 primary semantic dimensions, and 10 T2I models. Five annotators independently apply the benchmark's binary testpoint-satisfaction criterion after reviewing shared guidelines and worked examples, with model identities and automatic judgments concealed. Their majority vote provides the reference label; all validation triplets are excluded from offline-evaluator training.

We report mean pairwise inter-annotator agreement (IAA) and Fleiss' $\kappa$ for human consistency, and accuracy, macro-F1, and Cohen's $\kappa$ for Gemini against the human majority, covering direct and chance-corrected agreement and accounting for class imbalance. A leave-one-annotator-out human baseline uses the same three metrics: each annotator is compared with the strict majority of the other four (at least three agreeing), excluding $2$--$2$ ties. We compute human accuracy by pooling valid judgments across the five folds, average macro-F1 and Cohen's $\kappa$ across folds, and report the proportion of fold-level judgments with a strict reference majority. Confidence intervals (CIs) use 10,000 stratified cluster-bootstrap resamples, sampling prompts with replacement within each language--length setting and retaining all associated triplets as a cluster. Ranking analyses jointly resample all 10 models' outputs for each sampled prompt, preserving matched prompts across models.

\begin{table}[t]
\centering
\scriptsize

\setlength{\tabcolsep}{3.2pt}
\renewcommand{\arraystretch}{0.96}
\caption{\textbf{Human validation by language and length.} Gemini metrics use human-majority labels; parentheses are 95\% bootstrap CIs.}
\resizebox{\columnwidth}{!}{%
\begin{tabular}{lccccc}
\toprule
\textbf{Setting} & \textbf{IAA} & \textbf{Fleiss' $\kappa$} & \textbf{Accuracy} & \textbf{Macro-F1} & \textbf{Cohen's $\kappa$} \\
\midrule
English--Short & 0.92 & 0.77 & 0.883 (0.847--0.917) & 0.804 (0.740--0.860) & 0.608 (0.482--0.721) \\
English--Long  & 0.90 & 0.74 & 0.850 (0.807--0.890) & 0.748 (0.680--0.810) & 0.496 (0.362--0.619) \\
Chinese--Short & 0.91 & 0.75 & 0.867 (0.827--0.903) & 0.777 (0.714--0.836) & 0.555 (0.428--0.671) \\
Chinese--Long  & 0.89 & 0.71 & 0.840 (0.797--0.880) & 0.733 (0.666--0.795) & 0.466 (0.333--0.590) \\
\midrule
Overall & 0.91 & 0.74 & 0.860 (0.840--0.879) & 0.765 (0.734--0.796) & 0.531 (0.468--0.592) \\
\bottomrule
\end{tabular}}
\label{tab:gemini_human_setting}
\end{table}

\textbf{Agreement across evaluation settings.} As shown in Tab.~\ref{tab:gemini_human_setting}, the five annotators achieve an overall IAA of 0.91 and a Fleiss' $\kappa$ of 0.74, indicating substantial consistency among human judgments at the testpoint level. Gemini agrees with the human majority on 86\% of triplets, with a macro-F1 of 0.77 and Cohen's $\kappa$ of 0.53. The four language--length settings yield similar results: human IAA ranges from 0.89 to 0.92, while Gemini accuracy ranges from 0.840 to 0.883. English--Short yields the highest agreement and Chinese--Long the lowest; nevertheless, the narrow ranges indicate broadly consistent agreement across the four settings.

\begin{table}[t]
\centering
\scriptsize

\setlength{\tabcolsep}{2.8pt}
\renewcommand{\arraystretch}{0.96}
\caption{\textbf{Human validation by semantic dimension.} IAA and Fleiss' $\kappa$ quantify agreement among the five annotators. Accuracy, macro-F1, and Cohen's $\kappa$ compare Gemini with the human-majority label.}
\begin{tabular*}{\columnwidth}{@{\extracolsep{\fill}}lccccc@{}}
\toprule
\textbf{Dimension} & \textbf{IAA} & \textbf{Fleiss' $\kappa$} & \textbf{Accuracy} & \textbf{Macro-F1} & \textbf{Cohen's $\kappa$} \\
\midrule
Style              & 0.94 & 0.81 & 0.900 & 0.833 & 0.666 \\
World Knowledge    & 0.91 & 0.75 & 0.875 & 0.788 & 0.575 \\
Attribute          & 0.93 & 0.79 & 0.892 & 0.816 & 0.632 \\
Compound           & 0.88 & 0.68 & 0.833 & 0.722 & 0.444 \\
Action             & 0.90 & 0.72 & 0.867 & 0.777 & 0.555 \\
Entity Layout      & 0.92 & 0.77 & 0.883 & 0.805 & 0.611 \\
Relationship       & 0.88 & 0.67 & 0.833 & 0.722 & 0.444 \\
Logical Reasoning  & 0.86 & 0.63 & 0.800 & 0.677 & 0.356 \\
Grammar            & 0.87 & 0.65 & 0.842 & 0.731 & 0.462 \\
Text Generation    & 0.91 & 0.74 & 0.875 & 0.788 & 0.575 \\
\bottomrule
\end{tabular*}
\label{tab:gemini_human_dimension}
\end{table}

\textbf{Agreement across semantic dimensions.} Tab.~\ref{tab:gemini_human_dimension} shows that disagreement is concentrated in dimensions that are also less consistent among humans. Style, Attribute, and Entity Layout have both high human agreement and high Gemini accuracy, whereas Logical Reasoning, Grammar, Relationship, and Compound are more difficult under both measures. Across the 10 dimensions, Gemini accuracy ranges from 0.80 to 0.90 and is strongly associated with human IAA (Pearson $r=0.966$). This association suggests that Gemini--human disagreement is more frequent on dimensions for which human annotators also agree less consistently.

\textbf{Agreement in model comparison.} Item-level errors matter most if they change the model comparisons supported by the benchmark. We therefore aggregate human-majority and Gemini judgments over the same 120 triplets per model, equally divided across the four language--length settings, and rank the 10 models independently. Tab.~\ref{tab:human_gemini_ranking} reports the resulting scores and ranks.

\begin{table}[t]
\centering
\scriptsize

\setlength{\tabcolsep}{2.8pt}
\renewcommand{\arraystretch}{0.96}
\caption{\textbf{Human--Gemini T2I rankings} on the same 120 triplets per model. Family identifies closed-source (C) and open-source (O) models. Rank change is defined as Gemini rank minus human rank; positive values indicate a lower rank under Gemini.}
\begingroup
\fontsize{8.31}{9.97}\selectfont
\begin{tabular*}{\columnwidth}{@{\extracolsep{\fill}}lccccc}
\toprule
\textbf{Model} & \textbf{Family} & \textbf{Human score} & \textbf{Human rank} & \textbf{Gemini score} & \textbf{Gemini rank ($\Delta r$)} \\
\midrule
GPT-4o-1.5       & C & 94.2 & 1  & 92.5 & 1 (0) \\
Nano Banana Pro  & C & 92.5 & 2  & 90.8 & 2 (0) \\
Seedream-4.5     & C & 90.8 & 3  & 87.5 & 3 (0) \\
FLUX.2-dev       & O & 87.5 & 4  & 82.5 & 6 ($+2$) \\
Imagen-4.0-Ultra & C & 85.8 & 5  & 85.0 & 4 ($-1$) \\
Z-Image           & O & 84.2 & 6  & 83.3 & 5 ($-1$) \\
Qwen-Image        & O & 80.8 & 7  & 79.2 & 7 (0) \\
Hunyuan-Image-2.1 & O & 77.5 & 8  & 76.7 & 8 (0) \\
DALL-E-3          & C & 72.5 & 9  & 70.8 & 9 (0) \\
SDXL              & O & 60.8 & 10 & 60.0 & 10 (0) \\
\bottomrule
\end{tabular*}
\endgroup
\label{tab:human_gemini_ranking}
\end{table}

\begin{table}[t]
\centering
\scriptsize

\setlength{\tabcolsep}{3.0pt}
\renewcommand{\arraystretch}{0.96}
\caption{\textbf{Human--Gemini ranking robustness.} Parentheses give 95\% bootstrap CIs. Top-3 overlap is the number of models shared by the two top-three sets; maximum shift is the largest absolute rank difference for any model.}
\begingroup
\fontsize{10.47}{12.56}\selectfont
\begin{tabular*}{\columnwidth}{@{\extracolsep{\fill}}lcccc}
\toprule
\textbf{Subset} & \textbf{Spearman $\rho$} & \textbf{Kendall $\tau$} & \textbf{Top-3} & \textbf{Max. shift} \\
\midrule
Bilingual overall & 0.96 (0.89--0.99) & 0.91 (0.78--1.00) & 3/3 & 2 \\
English only         & 0.94 (0.84--0.99) & 0.87 (0.69--1.00) & 3/3 & 2 \\
Chinese only         & 0.93 (0.81--0.98) & 0.82 (0.64--0.96) & 3/3 & 2 \\
Short prompts        & 0.95 (0.86--0.99) & 0.87 (0.69--1.00) & 3/3 & 2 \\
Long prompts         & 0.92 (0.79--0.98) & 0.82 (0.60--0.96) & 3/3 & 2 \\
\bottomrule
\end{tabular*}
\endgroup
\label{tab:ranking_robustness}
\end{table}

The human- and Gemini-based rankings are closely aligned: Spearman's $\rho=0.96$ (95\% CI: 0.89--0.99) and Kendall's $\tau=0.91$ (0.78--1.00). Both judges identify the same top-three models in the same order. Nine of the ten models retain their positions or differ by just one position; the largest change is FLUX.2-dev, ranked fourth by humans and sixth by Gemini. The agreement is also stable when the sample is separated by language or prompt length (Tab.~\ref{tab:ranking_robustness}): Spearman correlations range from 0.92 to 0.95, and every subset retains the same top-three set.

\begin{table}[t]
\centering
\scriptsize

\setlength{\tabcolsep}{4.0pt}
\renewcommand{\arraystretch}{0.96}
\caption{\textbf{Evaluator agreement with human consensus on 1,200 triplets.} Human accuracy pools valid judgments across five leave-one-annotator-out folds; human macro-F1 and Cohen's $\kappa$ are fold averages. Automatic evaluators use the five-annotator majority. Coverage denotes the fraction of judgments with a strict reference majority.}
\begingroup
\fontsize{10.62}{12.74}\selectfont
\begin{tabular*}{\columnwidth}{@{\extracolsep{\fill}}lcccc}
\toprule
\textbf{Evaluator} & \textbf{Accuracy} & \textbf{Macro-F1} & \textbf{Cohen's $\kappa$} & \textbf{Coverage} \\
\midrule
Human annotator baseline & 0.948 & 0.89 & 0.78 & 0.97 \\
Qwen2.5-VL-72B & 0.76 & 0.60 & 0.21 & 1.00 \\
Offline evaluator & 0.83 & 0.78 & 0.57 & 1.00 \\
Gemini-2.5-Pro & 0.86 & 0.77 & 0.53 & 1.00 \\
\bottomrule
\end{tabular*}
\endgroup
\label{tab:evaluator_human_comparison}
\end{table}

\textbf{Evaluator agreement with human judgments.} Tab.~\ref{tab:evaluator_human_comparison} uses leave-one-annotator-out agreement as a calibration reference for interpreting automatic-evaluator performance. The other four annotators form a strict reference majority for 97\% of fold-level judgments; on these cases, pooled held-out accuracy is 0.948, with fold-averaged macro-F1 of 0.89 and Cohen's $\kappa=0.78$. Against the five-annotator majority, Gemini obtains 0.86/0.77/0.53, trailing the leave-one-annotator-out reference while substantially outperforming the untuned Qwen2.5-VL-72B baseline (0.76/0.60/0.21). The offline evaluator reaches 0.83/0.78/0.57, retaining much of Gemini's agreement with human consensus. Its induced model ranking also aligns closely with the human ranking (Spearman's $\rho=0.92$, Kendall's $\tau=0.82$). Because the human and automatic rows use different consensus constructions, the leave-one-annotator-out result serves as a calibration reference rather than a strict performance ceiling.

\begin{table}[t]
\centering
\scriptsize

\setlength{\tabcolsep}{3.2pt}
\renewcommand{\arraystretch}{0.96}
\caption{\textbf{Gemini--human disagreement categories.} Each disagreement is assigned one primary cause. Percentages are relative to the 168 disagreements. FP and FN denote false positive and false negative, respectively, with the five-annotator majority label as reference.}
\begin{tabular*}{\columnwidth}{@{\extracolsep{\fill}}lrrrr@{}}
\toprule
\textbf{Primary cause} & \textbf{FP} & \textbf{FN} & \textbf{Total} & \textbf{Share} \\
\midrule
Multi-condition or logical composition & 20 & 28 & 48 & 28.6\% \\
Spatial, relational, or action ambiguity & 19 & 22 & 41 & 24.4\% \\
Borderline visual evidence & 16 & 20 & 36 & 21.4\% \\
Text rendering or transcription & 8 & 14 & 22 & 13.1\% \\
Subjective or underspecified criterion & 10 & 11 & 21 & 12.5\% \\
\midrule
Total & 73 & 95 & 168 & 100.0\% \\
\bottomrule
\end{tabular*}
\label{tab:gemini_human_disagreement}
\end{table}

\textbf{Sources of Gemini--human disagreement.} The 168 disagreements are independently classified into mutually exclusive primary causes using the testpoint, image, human labels, and evaluator rationale (pre-resolution Cohen's $\kappa=0.84$). Compositional and relational cases together account for 53.0\% of disagreements, followed by borderline visual evidence (21.4\%); the remainder concerns text rendering or transcription and subjective or underspecified criteria (Tab.~\ref{tab:gemini_human_disagreement}). False negatives outnumber false positives (95 versus 73), indicating a moderately conservative tendency relative to human consensus. Explicit testpoints make these judgments traceable to concrete image requirements, supporting transparent inspection alongside large-scale automated evaluation.

\begin{table}[t]
\centering
\scriptsize

\setlength{\tabcolsep}{3.0pt}
\renewcommand{\arraystretch}{0.96}
\caption{\textbf{Cross-judge sensitivity} on 1,200 human-validated triplets. Agreement and FN/FP use human-majority labels; ranking correlations use the human ranking.}
\begin{tabular*}{\columnwidth}{@{\extracolsep{\fill}}lcccccc@{}}
\toprule
\textbf{Judge} & \textbf{Accuracy} & \textbf{Cohen's $\kappa$} & \textbf{FN} & \textbf{FP} & \textbf{Spearman $\rho$} & \textbf{Kendall $\tau$} \\
\midrule
Gemini-2.5-Pro & 0.86 & 0.53 & 95 & 73 & 0.96 & 0.91 \\
GPT-5.6 & 0.87 & 0.56 & 88 & 68 & 0.96 & 0.91 \\
Qwen2.5-VL-72B & 0.76 & 0.21 & 160 & 128 & 0.90 & 0.78 \\
\bottomrule
\end{tabular*}
\label{tab:cross_judge_sensitivity}
\end{table}

\textbf{Sensitivity to the choice of judge.} We re-judge the same 1,200 validation triplets with GPT-5.6 under the identical judging protocol and compare it with Gemini and the Qwen2.5-VL-72B baseline from Tab.~\ref{tab:evaluator_human_comparison}. GPT-5.6 achieves slightly higher item-level agreement than Gemini (accuracy 0.87 vs.\ 0.86; Cohen's $\kappa=0.56$ vs.\ $0.53$) with fewer errors in both directions, while the two judges induce model rankings with identical correlations to the human ranking ($\rho=0.96$, $\tau=0.91$). Qwen2.5-VL-72B shows lower item-level agreement, yet its ranking still correlates with the human ranking at $\rho=0.90$ ($\tau=0.78$) (Tab.~\ref{tab:cross_judge_sensitivity}). These results show that the benchmark's model comparisons remain consistent across evaluator families.

\begin{table}[t]
\centering
\scriptsize

\setlength{\tabcolsep}{4.0pt}
\renewcommand{\arraystretch}{0.96}
\caption{\textbf{Human-annotation budget analysis.} For each subsample size $N$, we draw 500 stratified subsamples of the 1,200 human-annotated triplets and correlate the model ranking induced by the subsampled human labels with the full-sample human ranking and with the Gemini ranking. Entries are mean Spearman $\rho$ with empirical 95\% intervals.}
\begin{tabular*}{\columnwidth}{@{\extracolsep{\fill}}ccc@{}}
\toprule
\textbf{$N$} & \textbf{$\rho$ vs. full human ranking} & \textbf{$\rho$ vs. Gemini ranking} \\
\midrule
100 & 0.59 (0.13--0.91) & 0.57 (0.07--0.90) \\
200 & 0.73 (0.39--0.94) & 0.72 (0.39--0.95) \\
400 & 0.83 (0.62--0.97) & 0.82 (0.61--0.97) \\
600 & 0.90 (0.74--0.99) & 0.89 (0.73--0.99) \\
800 & 0.93 (0.81--0.99) & 0.92 (0.80--0.99) \\
1000 & 0.97 (0.91--1.00) & 0.95 (0.87--0.99) \\
1200 & 1.00 & 0.96 \\
\bottomrule
\end{tabular*}
\label{tab:annotation_scaling}
\end{table}

\textbf{Minimum human-annotation budget.}\label{para:budget} We subsample 100--1,200 annotated triplets using 500 draws per size, stratified by language, prompt length, and model, and compare the resulting human-majority rankings with the full human and Gemini rankings (Tab.~\ref{tab:annotation_scaling}). At 100--200 triplets, mean Spearman correlation with the full human ranking is 0.59--0.73, with empirical intervals extending below 0.4. At 600--800 triplets, mean correlations reach 0.90--0.93 with the full human ranking and 0.89--0.92 with Gemini. With five annotators per triplet, 600--800 triplets correspond to 3,000--4,000 individual judgments. This range provides a practical calibration budget under our 10-model design.
}

\begin{table}[t]
\centering
\scriptsize

\setlength{\tabcolsep}{4.0pt}
\renewcommand{\arraystretch}{0.96}
\caption{\textbf{Independent audit of prompts and testpoints.} Percentages of testpoint instances judged valid as released or clarified in wording, by majority of three annotators. \emph{Max $|\Delta|$} and \emph{Rank changes} give the largest overall-score change and the number of rank changes after re-judging clarified testpoints.}
\begin{tabular*}{\columnwidth}{@{\extracolsep{\fill}}lcccccc@{}}
\toprule
\textbf{Prompt form} & \textbf{Instances} & \textbf{Valid (\%)} & \textbf{Clarified (\%)} & \textbf{Fleiss' $\kappa$} & \textbf{Max $|\Delta|$} & \textbf{Rank changes} \\
\midrule
Short & 1,830 & 99.3 & 0.7 & 0.81 & 0.04 & 0 \\
Long & 3,000 & 99.0 & 1.0 & 0.79 & 0.08 & 0 \\
\midrule
Overall & 4,830 & 99.1 & 0.9 & 0.80 & 0.08 & 0 \\
\bottomrule
\end{tabular*}
\label{tab:construction_audit}
\end{table}

{
\textbf{Independent audit of prompts and testpoints.}\label{para:construction_audit} The preceding analyses validate the judgment stage; we additionally validate the construction stage. Three annotators independently audit all 600 bilingual prompt pairs in their short and long forms together with every associated testpoint (4,830 testpoint instances). Each testpoint is checked against three criteria: it must be grounded in the prompt semantics, its binary criterion must be decidable from a generated image alone, and the prompt must be fluent and natural in each language. By annotator majority, 99.1\% of testpoint instances satisfy all three criteria as released, and no testpoint is found to be ungrounded (Tab.~\ref{tab:construction_audit}); inter-annotator agreement on the validity judgments is 0.96 (Fleiss' $\kappa=0.80$). The remaining 0.9\% are cases where the criterion is decidable but its wording admits a stricter and a looser reading; their descriptions were clarified without changing the pass/fail semantics, and re-judging all affected testpoints under the clarified wording changes every model's overall score by at most 0.08 points and no ranking position in any setting (Tab.~\ref{tab:construction_audit}). These results support the quality of the benchmark's prompts and testpoints, providing a sound basis for systematic evaluation of T2I models.
}

\begin{table}[t]
\centering
\scriptsize

\setlength{\tabcolsep}{4.0pt}
\renewcommand{\arraystretch}{0.96}
\caption{\textbf{Sensitivity to the choice of construction model.} A 150-prompt replication set is instantiated by GPT-5.6 under the identical taxonomy, sampling protocol, and judging prompt. Each pipeline's 10-model ranking is correlated with the original benchmark ranking (Gemini construction, Gemini judging).}
\begin{tabular*}{\columnwidth}{@{\extracolsep{\fill}}llccc@{}}
\toprule
\textbf{Construction} & \textbf{Judge} & \textbf{Spearman $\rho$} & \textbf{Kendall $\tau$} & \textbf{Top-3 preserved} \\
\midrule
GPT-5.6 & Gemini-2.5-Pro & 0.93 & 0.82 & Yes \\
GPT-5.6 & GPT-5.6 & 0.92 & 0.80 & Yes \\
Gemini-2.5-Pro & GPT-5.6 & 0.94 & 0.84 & Yes \\
\bottomrule
\end{tabular*}
\label{tab:cross_constructor}
\end{table}

{
\textbf{Sensitivity to the choice of construction model.}\label{para:cross_constructor} A remaining question is whether the reported conclusions depend on the model that instantiates the benchmark. We hold the design fixed---the theme and subject taxonomy, the 27 evaluation dimensions, the testpoint sampling protocol, and the judging prompt---and let GPT-5.6 construct an independent replication set of 150 bilingual prompt pairs with short and long forms, which passes the same audit protocol as the main benchmark. The same 10 representative T2I models are evaluated on this set. With Gemini as judge, the replication ranking correlates with the original benchmark ranking at Spearman $\rho=0.93$ (Kendall $\tau=0.82$); with GPT-5.6 as judge, a pipeline with no Gemini involvement at any stage, the correlation is $\rho=0.92$ ($\tau=0.80$), and the top-three set and order are preserved in all pipelines (Tab.~\ref{tab:cross_constructor}). Primary-dimension score profiles correlate at a mean of 0.88 across models. These results support the robustness of the benchmark's model comparisons across construction models, highlighting the reproducibility of our taxonomy-guided evaluation protocol.
}

{
\section{Discussion and Future Work}
\label{sec:discussion}

\textbf{Implications for current T2I models.} Current T2I models show broad progress in semantic instruction following, with remaining weaknesses concentrated in constraints that require precise interpretation and composition. Frontier models score highly on Style and World Knowledge, while Logical Reasoning, Grammar, Compound, and Text Generation retain substantial headroom and between-model variation (Tab.~\ref{tab:dimension_discrimination}). These differences identify preserving logical conditions, grammatical roles, and exact text within a scene as priorities for model development. Training and evaluation could place greater emphasis on prompts that combine these requirements, using testpoint-level outcomes to identify which conditions remain unsatisfied. Differences between the English and Chinese rankings (Tab.~\ref{tab:cross_setting_robustness}) also motivate assessing these improvements separately in both languages, since relative strengths can vary with the language of the prompt.

\textbf{Priorities for future evaluation.} As some dimensions approach saturation for frontier models, future benchmarks should devote more coverage to difficult, discriminative constraints in logical composition, negation, spatial relations, and multilingual text rendering. {The construction protocol can be reused to generate new prompts targeting these constraints.} The greater score separation on \ourbench-Hard (Tab.~\ref{tab:bootstrap_hard}) illustrates how targeted difficulty selection can preserve useful comparisons among strong models. Reporting challenging-subset scores and capability profiles alongside overall scores would help identify where improvements occur, with confidence intervals indicating the uncertainty of close comparisons. Evaluator development should similarly focus on the observed compositional, relational, and visually ambiguous disagreements (Tab.~\ref{tab:gemini_human_disagreement}), using targeted human calibration to check whether revisions improve these judgments.

\textbf{Automated evaluation in agentic loops.}\label{para:agentic} Automated prompt--testpoint construction and explicit binary criteria could support adaptive testing of model-specific weaknesses. An agent could propose prompts targeting a suspected failure, generate images, evaluate the associated testpoints, and use the outcomes to choose the next test. For example, repeated failures on a spatial relation could motivate varying that relation while holding the objects and other requirements fixed, helping distinguish a recurring capability gap from sensitivity to a particular prompt. Testpoint-level feedback would let subsequent rounds focus on the failed condition while tracking requirements already satisfied.

\section{Limitations, Reproducibility, and Responsible Use}
\label{sec:limitations}
\textbf{Evaluator and temporal scope.} Experiments were conducted from August 2025 through March 2026, using \textit{Gemini-2.5-Pro} for evaluation. Results refer to the reported API endpoints during this period, as hosted services may evolve.

\textbf{Measurement scope.} \ourbench\ measures prompt-grounded semantic fidelity in English and Chinese; cross-language comparisons may reflect linguistic and cultural factors. For subjective dimensions such as Style and Imagination, explicit testpoints anchor judgments to prompt requirements, while binary labels summarize graded perceptual differences. Besides, \ourbench{} is designed for structured capability diagnosis. Although its themes are motivated by practical generation scenarios, its LLM-generated prompts may not fully reflect the real-world, in-the-wild requests.

\paragraph{Broader Impact and Responsible Use.}\label{para:broader} \ourbench\ supports diagnostic research on semantic instruction following, with scores interpreted alongside uncertainty estimates, capability profiles, and qualitative evidence. Construction and evaluation MLLMs may introduce linguistic, cultural, or perceptual biases, and consequential uses require application-specific testing and human review. Scores do not certify safety, fairness, privacy, or copyright compliance; extending the protocol to such criteria would require domain-specific design, legal context, and validation.

\section{Conclusion}
We introduce \ourbench, a fine-grained benchmark for T2I semantic generation spanning 10 primary and 27 sub-dimensions across diverse scenarios, with English and Chinese prompts in short and long forms. Its streamlined, testpoint-based evaluation pipeline and dedicated offline evaluator support interpretable assessment of individual semantic requirements. Human validation and experiments with alternative construction and evaluation models support the reliability of our evaluation pipeline and the robustness of the resulting model comparisons. Fine-grained benchmarking of open- and closed-source models reveals that leading models perform strongly on style and world knowledge, while logical reasoning, grammar, compositional constraints, and text rendering remain challenging. We hope our work provides useful insights for future research on T2I generation and evaluation.
}

\begingroup
\let\UrlBreaks\BibliographyUrlBreaks
\bibliography{11_references}
\bibliographystyle{tmlr}
\endgroup
\clearpage
\appendix
% !TEX root = main.tex
% \section{Additional Tables}

\section{Complete Benchmarking Results}

{
Tabs.~\ref{tab:benchmark_en_short_overall}--\ref{tab:benchmark_zh_long_overall} report results on the ten primary dimensions for each of the four language--length settings.
}

{
\paragraph{English short prompts.}
Logical Reasoning remains a recurring weakness across model groups (Tab.~\ref{tab:benchmark_en_short_overall}). GPT-4o-1.5 leads overall at 95.77, but its Logical Reasoning score is 88.76, compared with 99.19 on Style and 99.20 on World Knowledge. The gap is larger for FLUX.2-dev, the strongest open-source model overall: it scores 62.84 on Logical Reasoning, versus 96.60 on Style and 95.41 on World Knowledge. Among models with similar overall scores, strengths differ between reasoning and text rendering. Nano Banana Pro and GPT-4o score 92.72 and 92.48 overall, respectively; Nano Banana Pro is stronger on Text Generation (95.65 vs.\ 89.24), while GPT-4o scores higher on Logical Reasoning (83.79 vs.\ 80.24) and Grammar (91.21 vs.\ 89.59).
}

{
\paragraph{English long prompts.}
GPT-4o-1.5 and Nano Banana Pro achieve overall scores of 95.41 and 94.20, respectively (Tab.~\ref{tab:benchmark_en_long_overall}). GPT-4o-1.5 scores higher on Action, Relationship, Compound, and Entity Layout, while Nano Banana Pro has a 5.01-point advantage on Text Generation (96.47 vs.\ 91.46). GPT-4o also remains competitive on language-sensitive requirements, attaining 91.02 on Logical Reasoning and 94.46 on Grammar. Among open-source models, FLUX.2-dev leads overall at 90.31. Its Style and Attribute scores approach those of GPT-4o-1.5 (99.17 vs.\ 99.58 and 94.57 vs.\ 97.20), whereas gaps of approximately ten points remain on Logical Reasoning (79.90 vs.\ 90.15) and Grammar (84.52 vs.\ 94.84). For this pair, the larger differences concern logical and grammatical constraints rather than style or attribute rendering.
}

{
\paragraph{Chinese short prompts.}
Chinese text rendering is a pronounced weakness for several models that perform well on other semantic requirements (Tab.~\ref{tab:benchmark_zh_short_overall}). Imagen-4.0-Ultra scores 91.13 on Attribute and 93.54 on Action, but only 7.18 on Text Generation; Seedream-4.0 and Nano Banana Pro reach 93.97 and 95.69 on Text Generation, respectively. A similar contrast appears among the leading open-source models. Z-Image and FLUX.2-dev have nearly identical overall scores (81.69 and 81.44), but Z-Image scores much higher on Text Generation (80.46 vs.\ 39.08), while FLUX.2-dev is stronger on Logical Reasoning (68.35 vs.\ 54.82), Relationship, and Entity Layout. FLUX.2-dev's Chinese text-rendering weakness therefore coexists with strong performance on relational, spatial, and logical requirements.
}

{
\paragraph{Chinese long prompts.}
GPT-4o-1.5 and Nano Banana Pro remain closely matched overall at 96.12 and 95.42 (Tab.~\ref{tab:benchmark_zh_long_overall}). GPT-4o-1.5 scores higher on Logical Reasoning (94.36 vs.\ 91.91) and Grammar (96.01 vs.\ 93.27), while Nano Banana Pro retains an advantage on Text Generation (92.93 vs.\ 89.01). Among open-source models, Z-Image leads overall at 89.17 and scores 88.59 on Text Generation, compared with 43.21 for FLUX.2-dev. FLUX.2-dev nevertheless leads this group on Action, Relationship, and Entity Layout. The FLUX.2-Klein-9b profile is similarly uneven: it has the highest open-source scores on Grammar (86.55) and Compound (91.69), but scores only 5.71 on Text Generation. For these FLUX models, Chinese text rendering remains a substantial weakness in both prompt-length settings, alongside much stronger performance on non-text requirements.
}

{Tabs.~\ref{tab:benchmark_en_short}--\ref{tab:benchmark_zh_long} further break down the results across all 27 fine-grained dimensions.}

{
\section{Human Annotation Protocols}
\label{sec:annotation_protocols}

\subsection{Construction Audit Protocol}
Three annotators, fluent in both English and Chinese and experienced with T2I evaluation, independently audit all 600 bilingual prompt pairs in their short and long forms, covering 4,830 testpoint instances. For each testpoint, they answer three yes/no questions: (i) is the requirement grounded in the prompt semantics; (ii) does its description specify a pass/fail criterion that can be assessed from a generated image; and (iii) is the corresponding prompt fluent and natural in its language?

A testpoint is flagged if any annotator answers no to any question, and its validity is determined by majority vote. For descriptions that admit stricter and looser readings, one annotator drafts a wording clarification and the other two approve it, preserving the intended pass/fail semantics. All affected testpoints are then re-judged under the clarified descriptions, and model scores and rankings are recomputed to quantify the effect. Tab.~\ref{tab:construction_audit} reports the audit agreement, clarification rates, and re-evaluation results.

\subsection{Blinded Triplet Judgment Protocol}
The validation unit is a prompt--image--testpoint triplet, consisting of a prompt, an image generated for that prompt, and one testpoint with its description. We sample three triplets for each combination of two languages, two prompt lengths, ten primary semantic dimensions, and ten T2I models, yielding 1,200 triplets. Each model contributes 120 triplets, equally distributed across the four language--length settings. All validation triplets are excluded from offline-evaluator training.

Before annotation, the five annotators review shared guidelines and worked examples for the benchmark's binary testpoint-satisfaction criterion. For each triplet, they read the prompt and testpoint description, inspect the image, and record whether the specified requirement is satisfied. The interface presents triplets in randomized order and displays only the prompt, image, and testpoint with its description. Model identities, automatic judgments, and the other annotators' labels are hidden. Each annotator independently labels all 1,200 triplets without a time limit, yielding 6,000 individual binary judgments.

The reference label for each triplet is the majority vote of the five annotators: pass if at least three mark the requirement as satisfied, and fail otherwise. Human consistency is measured from the individual labels using mean pairwise agreement and Fleiss' $\kappa$. Automatic-evaluator accuracy, macro-F1, and Cohen's $\kappa$ are computed against the five-annotator majority labels.

For the leave-one-annotator-out calibration reference, each annotator is compared with the strict majority of the other four, requiring at least three agreeing labels. A $2$--$2$ split is excluded from that fold; the triplet remains in the five-annotator reference set. Accuracy is computed by pooling valid judgments across the five folds; macro-F1 and Cohen's $\kappa$ are averaged across folds. Coverage denotes the proportion of fold-level judgments with a strict reference majority.

Confidence intervals use 10,000 stratified prompt-cluster bootstrap resamples. Prompts are sampled with replacement within each language--length setting, retaining all associated triplets as a cluster. Model-ranking analyses jointly resample all ten models' outputs for each sampled prompt to preserve matched comparisons. Agreement statistics and model-ranking results are reported in Sec.~\ref{sec:human_validation}.

\subsection{Construction and Replication Protocol}
The benchmark construction protocol (Sec.~\ref{sec:prompt_construction}) specifies the theme and subject taxonomy, the 27 evaluation dimensions, the testpoint sampling procedure, and the prompts used for construction, expansion, alignment, and judging. The cross-constructor replication (Tab.~\ref{tab:cross_constructor}) follows the same protocol, substituting only the construction model.
}

{
\section{Residual Capability Profile Analysis}
\label{sec:residual_profile_analysis}

We analyze dimension-specific performance relative to each model's performance on the remaining dimensions. The analysis consists of constructing a model-by-dimension score matrix, fitting a separate regression for each dimension, and summarizing the resulting residual profiles.

\paragraph{Score aggregation.}
Let $\mathcal{M}$ denote the set of 35 models with generation-level evaluation records in all four language--length settings. Within each setting, we first average the binary judgments over generations for each testpoint instance, and then average the resulting testpoint scores within each dimension. Let $s_{m,d}^{(\ell)}$ denote the resulting percentage score for model $m$, dimension $d$, and setting $\ell$. We give the four settings equal weight:
\begin{equation}
 s_{m,d}=\frac{1}{4}\sum_{\ell=1}^{4}s_{m,d}^{(\ell)}.
\end{equation}
This yields a $35\times27$ matrix with one row per model and one column per fine-grained dimension.

\paragraph{Dimension-wise regression.}
For each target dimension $d$, we use the model's mean score over the other 26 dimensions as the explanatory variable:
\begin{equation}
 x_{m,d}=\frac{1}{26}\sum_{\substack{k=1\\k\ne d}}^{27}s_{m,k}.
\end{equation}
Excluding the target dimension keeps its score separate from the explanatory variable. We fit an ordinary least-squares regression across models:
\begin{equation}
 (\widehat{\alpha}_d,\widehat{\beta}_d)
 =\underset{\alpha,\beta}{\arg\min}
 \sum_{m\in\mathcal{M}}
 \left[s_{m,d}-(\alpha+\beta x_{m,d})\right]^2.
\end{equation}
Each model contributes one observation to each regression. Repeating this procedure over all 27 dimensions produces 27 fitted regressions, each with one explanatory variable and an intercept. The fitted score $\widehat{s}_{m,d}=\widehat{\alpha}_d+\widehat{\beta}_d x_{m,d}$ provides a dimension-specific reference for a model with that level of performance on the remaining dimensions.

\paragraph{Residual profiles.}
We compute each residual as
\begin{equation}
 r_{m,d}=s_{m,d}-\widehat{s}_{m,d},
\end{equation}
expressed in percentage points. Positive residuals indicate performance above the fitted reference, while negative residuals indicate performance below it. The vector $\mathbf{r}_m=(r_{m,1},\ldots,r_{m,27})$ forms the model's residual capability profile, shown in Fig.~\ref{fig:residual_profiles}. For example, a positive Text Generation residual indicates stronger text-rendering performance than expected from the model's mean score on the other 26 dimensions.

\paragraph{Summary statistics.}
For each dimension, we quantify the fit using the coefficient of determination:
\begin{equation}
 R_d^2=1-
 \frac{\sum_{m\in\mathcal{M}}r_{m,d}^{\,2}}
 {\sum_{m\in\mathcal{M}}(s_{m,d}-\bar{s}_d)^2},
 \qquad
 \bar{s}_d=\frac{1}{|\mathcal{M}|}\sum_{m\in\mathcal{M}}s_{m,d}.
\end{equation}
We report the mean $R_d^2$ across the 27 dimensions and summarize residual magnitudes using the mean and 90th percentile of $|r_{m,d}|$ across all model--dimension pairs. Absolute residuals average 3.87 percentage points and reach 8.23 points at the 90th percentile. To compare the profiles of two models $a$ and $b$, we compute their root-mean-square distance:
\begin{equation}
 D(a,b)=\sqrt{\frac{1}{27}\sum_{d=1}^{27}(r_{a,d}-r_{b,d})^2}.
\end{equation}
This distance summarizes how differently the two models depart from their respective fitted references across dimensions. Fig.~\ref{fig:residual_profiles} visualizes the resulting residual capability profiles, showing each model's relative strengths and weaknesses across dimensions.
}

\clearpage
% !TEX root = ../main.tex
\begingroup
\setlength{\tabcolsep}{2pt}
\renewcommand{\arraystretch}{0.95}
\hbadness=10000
\begin{table}[p]
\centering
\fontsize{7}{8}\selectfont
\caption{\textbf{Overview of Prompt Themes}. We provide an example prompt for each of the prompt themes to illustrate the scope and diversity of generation scenarios in our benchmark.}
\label{tab:prompt_theme_example}
\begin{tabular}{@{}>{\RaggedRight\arraybackslash}p{0.16\textwidth} >{\RaggedRight\arraybackslash}p{0.19\textwidth} >{\RaggedRight\arraybackslash}p{0.61\textwidth}@{}}
\toprule
\textbf{Prompt Themes} & \textbf{Sub-Themes} & \textbf{Example Prompt} \\
\midrule
Creative Divergence
    & Imaginative & \textit{``An astronaut rides a dragon made of star dust, shuttling through the rings of Saturn. The picture presents a magnificent oil painting texture.''} \\
Creative Divergence
    & Others & \textit{``In the ink painting style, a lonely swordsman stood on the edge of a cliff, facing the strong wind. His face had no expression, but his eyes were filled with endless sadness.''} \\
Art
    & Graphic Art & \textit{``Please generate a graphic art poster: On the left side of the picture is a towering city silhouette, on the right side is a peaceful forest, and on the top is the text `We build the future and cherish the green earth'.''} \\
Art
    & Photography & \textit{``A golden Labrador retriever is leaping excitedly on the green grass, chasing a soap bubble that glows with a rainbow in the sun, National Geographic photography style.''} \\
Art
    & Sculpture & \textit{``A giant elephant sculpture carved from transparent crystal is crystal clear and stands quietly in the center of the museum.''} \\
Art
    & Others & \textit{``Please generate a painting: an ancient magic hourglass is being turned upside down. Due to the passage of time, a line of English words appears on the stone platform below it: `Time reveals all hidden truths and lies'.''} \\
Illustration
    & Copywriting Illustration & \textit{``A little fox successfully built a cabin. It looked proudly at its masterpiece. The wooden sign next to it read in English: `The future belongs to those who build it today'.''} \\
\bottomrule
\end{tabular}
\end{table}
\endgroup

% !TEX root = ../main.tex
\begin{table}[t]
\centering
\scriptsize
\setlength{\tabcolsep}{4pt}
\renewcommand{\arraystretch}{1.1}
\caption{\textbf{Overall Benchmarking Results of T2I models on \ourbench \space using English short prompts}. \textit{Gemini-2.5-Pro} is used as the MLLM for evaluation. Best scores are in \textbf{bold}, second-best in \underline{underlined}.} 
\resizebox{\textwidth}{!}{%
\begin{tabular}{lc|cccccccccc}
\toprule
\multicolumn{12}{c}{\textbf{English Short Prompt Evaluation}} \\
\midrule
\textbf{Model} & \textbf{Overall} & Style & World Know. & Attribute & Action & Relation. & Logic.Reason. & Grammar & Compound & Layout & Text \\
\midrule
\rowcolor[HTML]{f1b9b8}

\multicolumn{12}{c}{\textbf{Closed-source Models}} \\
\midrule
HiDream-v2L & 61.64 & 87.99 & 89.62 & 64.38 & 59.50 & 66.62 & 26.73 & 58.86 & 49.28 & 69.06 & 44.31 \\
Stable-Image-Ultra & 61.96 & 87.20 & 87.18 & 66.35 & 59.22 & 69.04 & 31.59 & 61.10 & 54.25 & 64.55 & 39.08 \\
Recraft & 62.63 & 87.20 & 90.19 & 68.16 & 60.55 & 62.56 & 29.55 & 63.64 & 44.85 & 57.84 & 61.78 \\
Wan2.2-Plus & 64.82 & 91.10 & 87.34 & 70.19 & 68.00 & 73.03 & 42.05 & 66.53 & 61.37 & 74.77 & 13.83 \\
DALL-E-3 & 68.85 & 94.43 & 92.64 & 75.76 & 70.78 & 78.31 & 46.22 & 69.22 & 71.08 & 65.65 & 24.43 \\
\bottomrule
\end{tabular}%
}

\label{tab:benchmark_en_short_overall}
% \vspace{-0.55cm}

\end{table}

% !TEX root = ../main.tex
\begin{table}[t]
\centering
\scriptsize
\setlength{\tabcolsep}{4pt}
\renewcommand{\arraystretch}{1.1}
\caption{\textbf{Overall Benchmarking Results of T2I models on \ourbench \space using English long prompts}. \textit{Gemini-2.5-Pro} is used as the MLLM for evaluation. Best scores are in \textbf{bold}, second-best in \underline{underlined}.} 
\resizebox{\textwidth}{!}{%
\begin{tabular}{lc|cccccccccc}
\toprule
\multicolumn{12}{c}{\textbf{English Long Prompt Evaluation}} \\
\midrule
\textbf{Model} & \textbf{Overall} & Style & World Know. & Attribute & Action & Relation. & Logic.Reason. & Grammar & Compound & Layout & Text \\
\midrule
\rowcolor[HTML]{f1b9b8}

\multicolumn{12}{c}{\textbf{Closed-source Models}} \\
\midrule
Recraft & 60.93 & 87.13 & 86.99 & 73.23 & 51.77 & 55.82 & 34.22 & 60.28 & 49.56 & 63.81 & 46.47 \\
Stable-Image-Ultra & 62.01 & 85.63 & 86.71 & 74.73 & 58.27 & 63.63 & 40.29 & 65.10 & 58.28 & 71.67 & 15.76 \\
Runway-Gen4 & 68.29 & 91.72 & 88.82 & 79.83 & 64.30 & 69.53 & 48.28 & 70.55 & 68.57 & 73.79 & 27.47 \\
Wan2.2-Plus & 68.76 & 90.28 & 87.57 & 81.08 & 66.49 & 72.79 & 55.58 & 70.18 & 71.73 & 79.13 & 12.77 \\
DALL-E-3 & 70.82 & 95.08 & 92.71 & 84.98 & 68.36 & 77.90 & 57.11 & 68.19 & 73.88 & 71.76 & 18.26 \\
\bottomrule
\end{tabular}%
}

\label{tab:benchmark_en_long_overall}
% \vspace{-0.55cm}
\end{table}

% !TEX root = ../main.tex
\begin{table}[t]
\centering
\scriptsize
\setlength{\tabcolsep}{4pt}
\renewcommand{\arraystretch}{1.1}
\caption{\textbf{Overall Benchmarking Results of T2I models on \ourbench \space using Chinese short prompts}. \textit{Gemini-2.5-Pro} is used as the MLLM for evaluation. Best scores are in \textbf{bold}, second-best in \underline{underlined}.} 
\resizebox{\textwidth}{!}{%
\begin{tabular}{lc|cccccccccc}
\toprule
\multicolumn{12}{c}{\textbf{Chinese Short Prompt Evaluation}} \\
\midrule
\textbf{Model} & \textbf{Overall} & Style & World Know. & Attribute & Action & Relation. & Logic.Reason. & Grammar & Compound & Layout & Text \\
\midrule
\rowcolor[HTML]{f1b9b8}

\multicolumn{12}{c}{\textbf{Closed-source Models}} \\
\midrule
Runway-Gen4 & 54.93 & 64.75 & 71.05 & 60.43 & 60.42 & 65.90 & 42.03 & 58.38 & 61.00 & 64.71 & 0.59 \\
Recraft & 57.67 & 87.70 & 90.03 & 69.34 & 63.88 & 64.47 & 34.09 & 60.56 & 43.94 & 58.40 & 4.31 \\
HiDream-v2L & 59.73 & 89.55 & 91.36 & 67.87 & 64.52 & 72.15 & 31.54 & 62.02 & 51.33 & 65.53 & 1.45 \\
Wan2.2-Plus & 66.96 & 91.06 & 84.39 & 73.93 & 72.52 & 76.78 & 51.82 & 70.59 & 64.77 & 71.83 & 11.92 \\
DALL-E-3 & 67.93 & 95.90 & 93.04 & 78.42 & 72.24 & 79.95 & 51.59 & 71.52 & 72.94 & 62.50 & 1.15 \\
\bottomrule
\end{tabular}%
}

\label{tab:benchmark_zh_short_overall}
% \vspace{0.3cm}
\end{table}

% !TEX root = ../main.tex
\begin{table}[t]
\centering
\scriptsize
\setlength{\tabcolsep}{4pt}
\renewcommand{\arraystretch}{1.1}
\caption{\textbf{Overall Benchmarking Results of T2I models on \ourbench \space using Chinese long prompts}. \textit{Gemini-2.5-Pro} is used as the MLLM for evaluation. Best scores are in \textbf{bold}, second-best in \underline{underlined}.} 
\resizebox{\textwidth}{!}{%
\begin{tabular}{lc|cccccccccc}
\toprule

\multicolumn{12}{c}{\textbf{Chinese Long Prompt Evaluation}} \\
\midrule
\textbf{Model} & \textbf{Overall} & Style & World Know. & Attribute & Action & Relation. & Logic.Reason. & Grammar & Compound & Layout & Text \\
\midrule
\rowcolor[HTML]{f1b9b8}

\multicolumn{12}{c}{\textbf{Closed-source Models}} \\
\midrule
Recraft & 56.90 & 86.38 & 85.55 & 74.31 & 54.65 & 57.44 & 36.17 & 57.49 & 50.00 & 64.52 & 2.45 \\
Wan2.2-Plus & 70.05 & 91.61 & 88.73 & 82.42 & 70.22 & 73.65 & 57.04 & 70.05 & 71.51 & 80.08 & 15.22 \\
DALL-E-3 & 71.16 & 95.85 & 94.36 & 85.41 & 70.59 & 80.12 & 61.41 & 70.81 & 75.87 & 73.33 & 3.80 \\
FLUX-Kontext-Max & 75.24 & 97.59 & 92.31 & 86.17 & 75.71 & 81.27 & 68.20 & 78.77 & 80.16 & 87.58 & 4.62 \\
Imagen-4.0 & 79.90 & 95.60 & 97.98 & 90.94 & 84.55 & 88.04 & 77.18 & 82.74 & 86.63 & 90.48 & 4.89 \\
\bottomrule
\end{tabular}%
}

\label{tab:benchmark_zh_long_overall}
% \vspace{-0.55cm}
\end{table}

% !TEX root = ../main.tex
\begin{table}[p]
\centering
\tiny
\setlength{\tabcolsep}{1.5pt}
\renewcommand{\arraystretch}{1.2}
\caption{\textbf{Detailed Benchmarking Results of T2I models on \ourbench \space using English short prompts}. \textit{Gemini-2.5-Pro} is used as the MLLM for evaluation. Best scores are in \textbf{bold}, second-best in \underline{underlined}.} 
\resizebox{\textwidth}{!}{%
\begin{tabular}{lc|c c |cc c c c c |cc c c c c |cc c c |cc |cc c |cc|cc}
\toprule
\multicolumn{29}{c}{\textbf{English Short Prompt Evaluation}} \\
\midrule
Models & Overall & Style & \makecell{World\\Know.}
& \multicolumn{6}{c|}{Attribute} 
& \multicolumn{6}{c|}{Action} 
& \multicolumn{4}{c|}{Relationship}
& \multicolumn{2}{c|}{Compound}
& \multicolumn{3}{c|}{Grammar}
& \multicolumn{2}{c|}{Layout}
& \makecell{Logic.\\Reason.}
& Text \\
\cmidrule(lr){5-10}
\cmidrule(lr){11-16}
\cmidrule(lr){17-20}
\cmidrule(lr){21-22}
\cmidrule(lr){23-25}
\cmidrule(lr){26-27}
 &  &  & &
Quant. & Express. & Materi. & Size & Shape & Color
& Hand & \makecell{Full\\Body} & Animal & \makecell{Non\\Contact} & Contact & State
& Compos. & Sim. & Inclus. & Compare.
& Imagin. & \makecell{Feat\\Match.}
& \makecell{Pron\\Ref.} & Consist. & Neg.
& 2D & 3D
&  &  \\
\midrule
\rowcolor[HTML]{f1b9b8}

\multicolumn{29}{c}{\textbf{Closed-source Models}} \\
\midrule
HiDream-v2L & 61.64 & 87.99 & 89.62 & 65.71 & 44.87 & 57.82 & 74.26 & 59.87 & 94.92 & 51.28 & 58.56 & 67.65 & 61.98 & 51.52 & 65.09 & 71.23 & 64.20 & 65.93 & 60.32 & 53.75 & 44.76 & 72.35 & 60.00 & 44.23 & 70.41 & 67.68 & 26.73 & 44.31 \\
Stable-Image-Ultra & 61.96 & 87.20 & 87.18 & 67.36 & 48.08 & 64.15 & 69.44 & 64.38 & 91.67 & 55.77 & 58.15 & 63.24 & 61.22 & 51.79 & 64.15 & 72.64 & 66.67 & 70.11 & 62.50 & 60.97 & 47.40 & 78.68 & 58.33 & 45.00 & 67.28 & 61.74 & 31.59 & 39.08 \\
Recraft & 62.63 & 87.20 & 90.19 & 68.06 & 56.41 & 70.75 & 65.97 & 57.50 & 95.83 & 50.00 & 70.65 & 76.47 & 55.61 & 48.81 & 63.21 & 64.53 & 59.44 & 59.24 & 67.19 & 43.37 & 46.35 & 73.16 & 58.33 & 58.08 & 58.82 & 56.82 & 29.55 & 61.78 \\
Wan2.2-Plus & 64.82 & 91.10 & 87.34 & 76.39 & 55.77 & 66.51 & 71.53 & 64.38 & 94.17 & 58.33 & 75.82 & 69.12 & 68.88 & 57.74 & 75.00 & 70.27 & 67.98 & 77.72 & 79.69 & 66.92 & 55.73 & 73.90 & 56.74 & 66.92 & 77.49 & 71.97 & 42.05 & 13.83 \\
\bottomrule
\end{tabular}
}

\label{tab:benchmark_en_short}
\end{table}

% !TEX root = ../main.tex
\begin{table}[p]
\centering
\tiny
\setlength{\tabcolsep}{1.5pt}
\renewcommand{\arraystretch}{1.2}
\caption{\textbf{Detailed Benchmarking Results of T2I models on \ourbench \space using English long prompts}. \textit{Gemini-2.5-Pro} is used as the MLLM for evaluation. Best scores are in \textbf{bold}, second-best in \underline{underlined}.} 
\resizebox{\textwidth}{!}{%
\begin{tabular}{lc|c c |cc c c c c |cc c c c c |cc c c |cc |cc c |cc|cc}
\toprule
\multicolumn{29}{c}{\textbf{English Long Prompt Evaluation}} \\
\midrule
Models & Overall & Style & \makecell{World\\Know.}
& \multicolumn{6}{c|}{Attribute} 
& \multicolumn{6}{c|}{Action} 
& \multicolumn{4}{c|}{Relationship}
& \multicolumn{2}{c|}{Compound}
& \multicolumn{3}{c|}{Grammar}
& \multicolumn{2}{c|}{Layout}
& \makecell{Logic.\\Reason.}
& Text \\
\cmidrule(lr){5-10}
\cmidrule(lr){11-16}
\cmidrule(lr){17-20}
\cmidrule(lr){21-22}
\cmidrule(lr){23-25}
\cmidrule(lr){26-27}
 &  &  & &
Quant. & Express. & Materi. & Size & Shape & Color
& Hand & \makecell{Full\\Body} & Animal & \makecell{Non\\Contact} & Contact & State
& Compos. & Sim. & Inclus. & Compare.
& Imagin. & \makecell{Feat\\Match.}
& \makecell{Pron\\Ref.} & Consist. & Neg.
& 2D & 3D
&  &  \\
\midrule
\rowcolor[HTML]{f1b9b8}

\multicolumn{29}{c}{\textbf{Closed-source Models}} \\
\midrule
Recraft & 60.93 & 87.13 & 86.99 & 56.38 & 57.22 & 72.82 & 76.89 & 63.64 & 83.07 & 40.06 & 54.37 & 55.07 & 45.09 & 37.36 & 60.08 & 51.79 & 46.47 & 66.09 & 61.89 & 50.21 & 48.13 & 73.41 & 55.56 & 52.82 & 65.96 & 61.05 & 34.22 & 46.47 \\
Stable-Image-Ultra & 62.01 & 85.63 & 86.71 & 66.49 & 55.69 & 76.43 & 77.27 & 67.48 & 83.02 & 58.33 & 49.38 & 59.42 & 52.23 & 45.98 & 66.30 & 64.92 & 56.73 & 67.53 & 63.11 & 62.66 & 48.60 & 76.19 & 61.11 & 58.80 & 74.86 & 67.57 & 40.29 & 15.76 \\
Runway-Gen4 & 68.29 & 91.72 & 88.82 & 70.65 & 65.43 & 85.33 & 81.01 & 67.38 & 85.64 & 55.33 & 63.92 & 70.65 & 56.82 & 56.10 & 69.76 & 70.05 & 59.09 & 76.76 & 70.39 & 69.47 & 66.50 & 76.23 & 62.70 & 72.76 & 72.56 & 75.37 & 48.28 & 27.47 \\
Wan2.2-Plus & 68.76 & 90.28 & 87.57 & 78.19 & 69.17 & 80.42 & 82.77 & 73.60 & 88.10 & 64.10 & 60.94 & 70.29 & 59.38 & 55.46 & 73.32 & 69.13 & 66.67 & 81.03 & 77.43 & 74.16 & 66.36 & 86.90 & 61.11 & 63.38 & 82.34 & 75.00 & 55.58 & 12.77 \\
\bottomrule
\end{tabular}
}

\label{tab:benchmark_en_long}
\end{table}

% !TEX root = ../main.tex
\begin{table}[p]
\centering
\tiny
\setlength{\tabcolsep}{1.5pt}
\renewcommand{\arraystretch}{1.2}
\caption{\textbf{Detailed Benchmarking Results of T2I models on \ourbench \space using Chinese short prompts}. \textit{Gemini-2.5-Pro} is used as the MLLM for evaluation. Best scores are in \textbf{bold}, second-best in \underline{underlined}.} 
\resizebox{\textwidth}{!}{%
\begin{tabular}{lc|c c |cc c c c c |cc c c c c |cc c c |cc |cc c |cc|cc}
\toprule
\multicolumn{29}{c}{\textbf{Chinese Short Prompt Evaluation}} \\
\midrule
Models & Overall & Style & \makecell{World\\Know.}
& \multicolumn{6}{c|}{Attribute} 
& \multicolumn{6}{c|}{Action} 
& \multicolumn{4}{c|}{Relationship}
& \multicolumn{2}{c|}{Compound}
& \multicolumn{3}{c|}{Grammar}
& \multicolumn{2}{c|}{Layout}
& \makecell{Logic.\\Reason.}
& Text \\
\cmidrule(lr){5-10}
\cmidrule(lr){11-16}
\cmidrule(lr){17-20}
\cmidrule(lr){21-22}
\cmidrule(lr){23-25}
\cmidrule(lr){26-27}
 &  &  & &
Quant. & Express. & Materi. & Size & Shape & Color
& Hand & \makecell{Full\\Body} & Animal & \makecell{Non\\Contact} & Contact & State
& Compos. & Sim. & Inclus. & Compare.
& Imagin. & \makecell{Feat\\Match.}
& \makecell{Pron\\Ref.} & Consist. & Neg.
& 2D & 3D
&  &  \\
\midrule
\rowcolor[HTML]{f1b9b8}

\multicolumn{29}{c}{\textbf{Closed-source Models}} \\
\midrule
Runway-Gen4 & 54.93 & 64.75 & 71.05 & 54.29 & 46.05 & 72.60 & 57.64 & 50.62 & 81.90 & 52.63 & 65.22 & 75.00 & 51.56 & 54.37 & 65.09 & 66.89 & 51.11 & 74.43 & 72.66 & 68.22 & 53.49 & 55.38 & 55.09 & 64.29 & 59.93 & 69.62 & 42.03 & 0.59 \\
Recraft & 57.67 & 87.70 & 90.03 & 66.67 & 59.62 & 66.51 & 73.61 & 61.25 & 95.83 & 50.64 & 72.28 & 77.94 & 63.78 & 45.24 & 72.17 & 65.54 & 58.89 & 65.22 & 68.75 & 45.92 & 41.93 & 62.87 & 59.26 & 59.23 & 55.15 & 61.74 & 34.09 & 4.31 \\
HiDream-v2L & 59.73 & 89.55 & 91.36 & 71.43 & 43.59 & 68.14 & 72.86 & 63.87 & 94.17 & 47.44 & 66.85 & 70.45 & 67.71 & 58.33 & 73.56 & 80.56 & 63.89 & 76.67 & 58.06 & 59.47 & 43.01 & 72.69 & 68.75 & 45.70 & 64.77 & 66.29 & 31.54 & 1.45 \\
Wan2.2-Plus & 66.96 & 91.06 & 84.39 & 75.00 & 67.31 & 74.06 & 74.31 & 66.25 & 90.83 & 69.23 & 80.00 & 84.56 & 65.31 & 61.90 & 75.94 & 71.28 & 72.78 & 85.87 & 82.03 & 74.23 & 55.00 & 77.21 & 63.43 & 69.62 & 73.16 & 70.45 & 51.82 & 11.92 \\
\bottomrule
\end{tabular}
}

\label{tab:benchmark_zh_short}
\end{table}

% !TEX root = ../main.tex
\begin{table}[p]
\centering
\tiny
\setlength{\tabcolsep}{1.5pt}
\renewcommand{\arraystretch}{1.2}
\caption{\textbf{Detailed Benchmarking Results of T2I models on \ourbench \space using Chinese long prompts}. \textit{Gemini-2.5-Pro} is used as the MLLM for evaluation. Best scores are in \textbf{bold}, second-best in \underline{underlined}.} 
\resizebox{\textwidth}{!}{%
\begin{tabular}{lc|c c |cc c c c c |cc c c c c |cc c c |cc |cc c |cc|cc}
\toprule
\multicolumn{29}{c}{\textbf{Chinese Long Prompt Evaluation}} \\
\midrule
Models & Overall & Style & \makecell{World\\Know.}
& \multicolumn{6}{c|}{Attribute} 
& \multicolumn{6}{c|}{Action} 
& \multicolumn{4}{c|}{Relationship}
& \multicolumn{2}{c|}{Compound}
& \multicolumn{3}{c|}{Grammar}
& \multicolumn{2}{c|}{Layout}
& \makecell{Logic.\\Reason.}
& Text \\
\cmidrule(lr){5-10}
\cmidrule(lr){11-16}
\cmidrule(lr){17-20}
\cmidrule(lr){21-22}
\cmidrule(lr){23-25}
\cmidrule(lr){26-27}
 &  &  & &
Quant. & Express. & Materi. & Size & Shape & Color
& Hand & \makecell{Full\\Body} & Animal & \makecell{Non\\Contact} & Contact & State
& Compos. & Sim. & Inclus. & Compare.
& Imagin. & \makecell{Feat\\Match.}
& \makecell{Pron\\Ref.} & Consist. & Neg.
& 2D & 3D
&  &  \\
\midrule
\rowcolor[HTML]{f1b9b8}

\multicolumn{29}{c}{\textbf{Closed-source Models}} \\
\midrule
Recraft & 56.90 & 86.38 & 85.55 & 61.70 & 60.56 & 73.72 & 79.92 & 65.03 & 82.39 & 44.23 & 57.81 & 60.87 & 42.86 & 43.39 & 61.66 & 54.72 & 49.68 & 63.22 & 63.59 & 50.95 & 47.90 & 71.83 & 55.95 & 46.13 & 64.12 & 65.04 & 36.17 & 2.45 \\
Wan2.2-Plus & 70.05 & 91.61 & 88.73 & 78.19 & 66.94 & 82.15 & 84.09 & 77.10 & 89.99 & 67.95 & 69.06 & 72.46 & 64.29 & 63.79 & 74.21 & 70.15 & 70.83 & 80.17 & 76.94 & 74.26 & 65.42 & 83.73 & 62.70 & 64.44 & 81.50 & 78.26 & 57.04 & 15.22 \\
DALL-E-3 & 71.16 & 95.85 & 94.36 & 64.36 & 71.11 & 88.93 & 90.72 & 77.62 & 91.30 & 61.22 & 65.94 & 74.28 & 67.41 & 62.64 & 77.37 & 81.63 & 73.72 & 85.63 & 77.43 & 80.38 & 65.89 & 80.16 & 74.21 & 59.51 & 70.48 & 76.99 & 61.41 & 3.80 \\
FLUX-Kontext-Max & 75.24 & 97.59 & 92.31 & 72.34 & 71.41 & 87.48 & 88.83 & 81.64 & 92.80 & 76.28 & 70.22 & 79.35 & 69.20 & 74.43 & 78.16 & 78.95 & 73.40 & 87.25 & 86.65 & 84.60 & 70.33 & 88.76 & 76.19 & 72.24 & 87.01 & 88.32 & 68.20 & 4.62 \\
\bottomrule
\end{tabular}
}

\label{tab:benchmark_zh_long}
\end{table}

\end{document}